%% file: main.tex
\documentclass{article}

\usepackage[nonatbib, preprint]{neurips_2020}

\usepackage[utf8]{inputenc} 
\usepackage[T1]{fontenc}    
\usepackage{hyperref}       
\usepackage{url}            
\usepackage{booktabs}       
\usepackage{amsfonts}       
\usepackage{nicefrac}       
\usepackage{microtype}      

\usepackage[english]{babel}
\usepackage{ifpdf}
\usepackage{cite} 
\usepackage{color}
\usepackage{pgf, tikz, pgfplots}
\usetikzlibrary{shapes, arrows, automata, plotmarks}
\usetikzlibrary{calc,hobby,decorations}
\usepackage[cmex10]{amsmath}
\usepackage{amssymb, amsthm}
\usepackage{mathrsfs}
\usepackage{algorithm,algpseudocode}
\algnewcommand{\LeftComment}[1]{\Statex \(\triangleright\) #1}
\usepackage{enumerate}
\usepackage{multirow}
\usepackage{rotating}
\usepackage{subcaption}
\input{mySections.tex}

\input{mySymbol.sty}
\input{pennColors.sty}

\def\Tr{\mathsf{T}}

\newtheorem{lemma}{\hspace{0pt}\bf Lemma}
\newtheorem{proposition}{\hspace{0pt}\bf Proposition}

\newtheorem{theorem}{\hspace{0pt}\bf Theorem}

\newtheorem{remark}{\hspace{0pt}\bf Remark}

\newtheorem{definition}{\hspace{0pt}\bf Definition}

\title{Training Robust Graph Neural Networks with Topology Adaptive Edge Dropping}

\author{%
   Zhan Gao$^{1}$ ~ Subhrajit Bhattacharya$^{2}$ ~ Leiming Zhang$^{2}$\\
   \AND
   Rick S. Blum$^{3}$ ~ Alejandro Ribeiro$^{1}$ ~ Brian M. Sadler$^{4}$\\
   \\
  $^{1}$ Department of Electrical and Systems Engineering, University of Pennsylvania, USA\\
  $^{2}$ Department of Mechanical Engineering and Mechanics, Lehigh University, USA \\
  $^{3}$ Department of Electrical and Computer Engineering, Lehigh University, USA \\ 
  $^{4}$ Army Research Laboratory, USA
}

\begin{document}

\maketitle

\begin{abstract}

Graph neural networks (GNNs) are processing architectures that exploit graph structural information to model representations from network data. Despite their success, GNNs suffer from sub-optimal generalization performance given limited training data, referred to as over-fitting. This paper proposes Topology Adaptive Edge Dropping (TADropEdge) method as an adaptive data augmentation technique to improve generalization performance and learn robust GNN models. We start by explicitly analyzing how random edge dropping increases the data diversity during training, while indicating i.i.d. edge dropping does not account for graph structural information and could result in noisy augmented data degrading performance. To overcome this issue, we consider graph connectivity as the key property that captures graph topology. TADropEdge incorporates this factor into random edge dropping such that the edge-dropped subgraphs maintain similar topology as the underlying graph, yielding more satisfactory data augmentation. In particular, TADropEdge first leverages the graph spectrum to assign proper weights to graph edges, which represent their criticality for establishing the graph connectivity. It then normalizes the edge weights and drops graph edges adaptively based on their normalized weights. Besides improving generalization performance, TADropEdge reduces variance for efficient training and can be applied as a generic method modular to different GNN models. Intensive experiments on real-life and synthetic datasets corroborate theory and verify the effectiveness of the proposed method.

\end{abstract}\vspace{-3mm}

\section{Introduction}

Modern problems of interest involve big data generated from social \cite{freeman2000visualizing}, citation \cite{zhang2018name}, biological \cite{fout2017protein} and physical networks \cite{liang2014wireless}, which can be modeled as graph signals. Analyzing graph signals requires information processing architectures that adapt to the irregular structure inherent in the underlying graph. Graph neural networks (GNNs) have been proposed as one such example, which leverage graph structural information to learn meaningful representations from graph signals \cite{Bruna2013, Defferrard2016, Fernando2019, Wu19-SGC, Kipf2017, Xu19-GIN, gori2005new, scarselli2008graph, battaglia2016interaction, gilmer2017neural, velivckovic2017graph, lee2018graph, wu2019dual} and perform successfully in a wide array of applications, e.g., node classification \cite{bhagat2011node, zhang2018end}, social recommendation \cite{Ying2018, Wu20192}, resource allocation \cite{zhan2020resource}, and planning among others \cite{yu2018learning, gao2021wide, li2018diffusion}.

While achieving resounding success, the performance of GNNs relies on the quality of training data. Given limited training data, GNNs learn over-parametric models that fit too closely to a particular dataset but fail to fit unseen testing data. The latter results in sub-optimal generalization performance, referred to as over-fitting \cite{dietterich1995overfitting}. Dropout is a conventional method to tackle this issue \cite{srivastava2014dropout}, which randomly omits a set of units from neural networks during training to reduce the number of parameters to be learned. In analogy to Dropout, multiple recent works focus on sampling the underlying graph to alleviate over-fitting. The works in \cite{zheng2020robust, luo2021learning} learn task-irrelevant graph edges with deep neural networks and prune away these edges to avoid aggregating unnecessary information. As these approaches require to learn the sampling parameter for each edge, they may be computationally expensive and tie to specific task-cases. On the other hand, the works in \cite{hamilton2017inductive, chen2018fastgcn, huang2018adaptive} have developed fast layer sampling methods, including neighbor-level sampling \cite{hamilton2017inductive}, node-level sampling \cite{chen2018fastgcn} and a layer-dependent variant \cite{huang2018adaptive}. While speeding up training, they may face "neighbor explosion" as layer-wise methods. DropEdge in \cite{rong2019dropedge} recently has been developed as a graph sampling method. It randomly removes a set of graph edges during training to prevent over-smoothing \cite{li2018deeper}. By producing varying graph connections, it could work as an intuitive data augmentation technique for over-fitting.

However, DropEdge considers each edge with equal priority and samples edges in an i.i.d. manner. Such a procedure does not account for the underlying graph structure and may yield edge-dropped subgraphs substantially different from the underlying one. These subgraphs then carry little structural information embedded in graph signals, and GNNs built upon these subgraphs during training could result in noisy information that degrades performance. From this perspective, we hypothesize that the structural information of the underlying graph should be incorporated during random edge dropping to avoid potential noise that hurts the training process.

This paper proposes TADropEdge (Topology Adaptive Edge Dropping) method as an adaptive data augmentation technique, which samples edge-dropped subgraphs during training while maintaining the overall topology of the underlying graph. 
We begin by explicitly analyzing how random edge dropping increases the data diversity alleviating over-fitting, and theoretically explain why i.i.d. edge dropping results in noisy augmented data degrading performance. We proceed to consider graph connectivity as the key property that captures graph topology, and develop TADropEdge that incorporates this factor into random edge dropping, which consists of three phases: i) Preprocess the underlying graph to identify disjoint components; ii) Leverage the graph spectrum to assign connectivity-relevant edge weights for each disjoint component; iii) Normalize the edge weights and drop graph edges adaptively based on their normalized weights. For the edge weight computation, we follow \cite{zhan2021barrier-journal} to employ \emph{aggregate resistance weights}, which represent the criticality of edges to the component connectivity. The latter is a positive real-valued distribution over edges that can be conveniently used for the adaptive edge dropping, captures graph topology more accurately, and is computationally more efficient for large-scale graphs, compared to some other methods \cite{dong2012clustering, hagen1992new, cheng2017network}.

To sum up, we make the following contributions: \textbf{I.} We provide explicit theoretical analysis w.r.t. how random edge dropping augments training data and why an i.i.d. procedure is not satisfactory. \textbf{II.} We propose TADropEdge that accounts for graph structural information during random edge dropping, yielding better data augmentation and improved generalization performance. \textbf{III.} TADropEdge has reduced variance by neglecting irrelevant noisy subgraphs for mitigating the training difficulty, and works as a generic method modular to different GNN models. Experimental results on real-life and synthetic datasets corroborate theoretical findings and demonstrate that TADropEdge effectively improves generalization performance and learns robust GNN models. Note that proofs, implementation details and more experimental results can be found in the supplementary material. \vspace{-2mm}

\section{Related Work}

\textbf{Graph neural networks.} GNNs have been developed as powerful tools that are capable of leveraging graph structural information to process graph signals. The most popular model for GNNs is the one involving graph convolutions, referred to as graph convolutional neural networks (GCNNs). Inspired by CNNs in the Euclidean domain, GCNNs consist of a cascade of layers, each of which applies a graph convolution \cite{Ortega18-GSP} followed by a pointwise nonlinearity. Several implementations have been developed under this framework \cite{Bruna2013, Defferrard2016, Fernando2019, Wu19-SGC, Kipf2017, Xu19-GIN}. The work in \cite{Bruna2013} computes the graph convolution in the spectral domain, while the authors in \cite{Defferrard2016} use a Chebyshev polynomial implementation. In parallel, the works in \cite{Fernando2019, Wu19-SGC} employ a summation polynomial to implement the graph convolution, and the authors in \cite{Kipf2017, Xu19-GIN} reduce the polynomial to the first order. In addition to GCNNs, other GNN models include message passing neural networks \cite{gori2005new, scarselli2008graph, battaglia2016interaction, gilmer2017neural} and graph attention networks \cite{velivckovic2017graph, lee2018graph, wu2019dual}. The former leverage the graph structure as the computation graph and combine arbitrary information across edges, while the latter learn the edge weights from training data as well. We in particular focus on GCNNs, while the proposed TADropEdge is applicable for any GNN models.

\textbf{Graph Sampling.} Graph sampling has been investigated in GNNs for efficient computation and performance enhancement. GraphSAGE in \cite{hamilton2017inductive} randomly samples a fixed-size neighborhood of each node and aggregates features accordingly for information fusion. FastGCN in \cite{chen2018fastgcn} interprets graph convolutions as integral transforms under probability measures and independently samples nodes at each layer. The work in \cite{huang2018adaptive} improves FastGCN by sampling nodes in lower layers conditioned on ones in upper layers. Instead of sampling layers, GraphSAINT in \cite{zeng2019graphsaint} samples the training graph and builds full GCNNs on the sampled subgraphs to avoid "neighbor explosion". DropEdge in \cite{rong2019dropedge} similarly samples the training graph, but focuses on dropping edges rather than nodes for alleviating over-smoothing. More recent work in \cite{gao2020stochastic} considers distributed scenarios over physical networks, where random edge dropping arises naturally during testing due to external factors. The authors in \cite{gao2020stochastic} developed SGNNs to incorporate such graph randomness into training and enhance robustness to the neighborhood uncertainty. However, the aforementioned works randomly sample nodes / edges during training without accounting for graph structural information inherent in graph signals, which is the key reason behind the success of GNNs. The proposed TADropEdge takes this factor into consideration and performs random edge dropping based on the underlying graph topology. \vspace{-2mm}

\section{Preliminaries}\label{sec:preliminary}

\textbf{GCNNs.} Let $\ccalG=(\ccalV, \ccalE)$ be a graph with the node set $\ccalV  = \{ n_i \}_{i=1}^N$ and the edge set $\ccalE=\{ e_m = (n_{i_m},n_{j_m}) \}_{m=1}^M$. The graph signal $\bbX: \ccalV \to \mathbb{R}^F$ is defined on the top of nodes, which assigns an $F$-dimensional feature to each node. The graph shift operator $\bbS \in \mathbb{R}^{N \times N}$ is a support matrix that satisfies $[\bbS]_{ij} \ne 0$ if $i = j$ or $(n_i,n_j)\in \ccalE$ and $[\bbS]_{ij}=0$ otherwise, thus captures the graph structure. Common examples of $\bbS$ include the adjacency, the Laplacian and their normalized versions. 

Graph convolutional neural networks (GCNNs) leverage graph structure to model nonlinear representations from graph signals. The GCNN is a cascade of layers, each layer consisting of a \emph{graph convolutional filter} followed by a pointwise nonlinearity. At layer $\ell$, we have an $F_{\ell-1}$-dimensional input signal $\bbX_{\ell-1} \in \mathbb{R}^{N \times F_{\ell-1}}$. The graph convolutional filter is a linear mapping of graph signals $\bbH_\ell: \mathbb{R}^{N \times F_{\ell-1}} \to \mathbb{R}^{N \times F_\ell}$, which is a polynomial function of the graph shift operator \cite{Fernando2019, Ortega18-GSP}\vspace{-2mm}
\begin{align} \label{eq:graphFilter}
\bbH_\ell(\bbS) \bbX_{\ell-1} := \sum_{k=0}^K \bbS^k \bbX_{\ell-1} \bbB_{\ell,k}
\end{align}
where $\ccalB_\ell = \{ \bbB_{\ell,k} \in \mathbb{R}^{F_{\ell-1} \times F_\ell} \}_{k=0}^K$ are filter parameters. The graph shift operation $\bbS \bbX$ represents the information exchange between neighboring nodes, and $\bbS^k \bbX$ accesses farther nodes in a $k$-hop neighborhood. Therefore, the graph filter is a shift-and-sum operation that aggregates the neighborhood information up to a radius of $K$. The feature generated by the filter is then passed through a pointwise nonlinearity $\sigma(\cdot): \mathbb{R} \to \mathbb{R}$ to produce the $\ell$th layer output feature\vspace{-1mm}
\begin{align} \label{eq:layerProcessing}
\bbX_\ell = \sigma\big( \bbH_\ell (\bbS) \bbX_{\ell-1} \big),~\forall~ \ell=1, \ldots, L.
\end{align}
The input to the GCNN is the $0$th layer signal $\bbX_0 = \bbX \in \mathbb{R}^{N \times F}$ and the output of the GCNN is the $L$th layer feature $\bbX_L \in \mathbb{R}^{N \times F_{L}}$. We interpret the GCNN as a nonlinear mapping of graph signals $\bbPhi(\bbX; \bbS, \ccalA):  \mathbb{R}^{N \times F} \to \mathbb{R}^{N \times F_L}$, where $\ccalA = \{ \ccalB_\ell \}_{\ell=1}^L$ are architecture parameters. If particularizing the filter order $K=1$, the GCNN reduces to the graph convolutional network (GCN) \cite{Kipf2017}.

\textbf{DropEdge.} DropEdge is a training technique that prevents over-smoothing and over-fitting for deep GCNNs \cite{rong2019dropedge}. At each training epoch $t$, DropEdge samples graph edges independently with the same default probability $p$ to obtain a sparse subgraph $\ccalG_t$ associated to the graph shift operator $\bbS_t$. It then replaces $\bbS$ with $\bbS_t$ in the GCNN architecture \eqref{eq:layerProcessing} for signal propagation and parameter training. In the validation and testing phases, the underlying graph $\ccalG$ is utilized without DropEdge.

In what follows, we first provide theoretical analysis on random edge dropping w.r.t. data augmentation. This analysis shows explicitly how it endows GCNNs with enhanced robustness to unseen signals, while indicating an i.i.d. dropping could result in noisy augmented data degrading performance. We then develop TADropEdge based on these theoretical findings, which accounts for graph structural information during random edge dropping and yields more satisfactory data augmentation.\vspace{-1mm}

\section{Analysis and Motivation} \label{sec:Motivation}\vspace{-0.5mm}

Given the training data $\ccalT = \{ (\bbX_r, \bby_r) \}_{\tau=1}^R$ of $R$ signal-label pairs and the loss function $c(\bbPhi(\bbX_r; \bbS, \ccalA), \bby_r)$ between the output feature and the label, we are interested in the cost over $\ccalT$ given by $C(\bbS, \ccalA) = 1/R \sum_{r=1}^R c(\bbPhi(\bbX_r; \bbS, \ccalA), \bby_r)$. 
Since the edge dropping incorporates randomness into training, the cost $C(\bbS_t, \ccalA)$ is a random variable depending on the edge-dropped subgraph $\bbS_t$ at training epoch $t$. The latter is sampled from an i.i.d. distribution $m_p(\bbS)$ determined by the default probability $p$. This observation motivates to consider the cost average over the graph distribution $m_p(\bbS)$, which formulates the stochastic optimization problem as \vspace{-1mm}
\begin{equation} \label{eq:stochasticProblemOverS}
\begin{split}
\min_{\ccalA} \bar{C}(\bbS, \ccalA) = \min_{\ccalA} \mathbb{E}_{\bbS \backsim m_p(\bbS)} \left[C({\bbS}, \ccalA)\right]. 
\end{split}
\end{equation}
The problem \eqref{eq:stochasticProblemOverS} is akin to the conventional stochastic optimization problem, while the expectation $\mathbb{E}[\cdot]$ is now w.r.t. graph randomness rather than data distribution. Since the cost $C(\bbS_t, \ccalA)$ is entirely determined by the subgraph $\bbS_t$, sampling $\bbS_t$ from $m_p(\bbS)$ is equivalent to sampling $C(\bbS_t, \ccalA)$ from $\mathbb{E}_{\bbS \backsim m_p(\bbS)} \left[C({\bbS}, \ccalA)\right]$ at training epoch $t$. Therefore, we can interpret DropEdge as running stochastic gradient descent (SGD) on \eqref{eq:stochasticProblemOverS}. We then formally analyze how DropEdge arguments the training data.

\subsection{Data augmentation}

We conduct analysis in the graph spectral domain and consider a one-dimensional graph signal $\bbx \in \mathbb{R}^{N}$ without loss of generality. Since the graph shift operator $\bbS$ is symmetric, it allows for the eigendecomposition $\bbS = \bbV \bbLambda \bbV^\top$ with orthogonal eigenvectors $\bbV \!=\! [\bbv_1,...,\bbv_N]$ and eigenvalues $\bbLambda \!=\! \text{diag} (\lambda_1,...,\lambda_N)$. We define the graph Fourier transform (GFT) by projecting $\bbx$ on the eigenvector basis as $\hat{\bbx} \!=\! \bbV^\top\! \bbx$. Substituting the GFT into the graph filter \eqref{eq:graphFilter} and using $\bbS^k = \bbV \bbLambda^k \bbV^\top$ yields \vspace{-2mm}
\begin{equation}\label{eq:FilterGFT}
\hat{\bbu} = \bbV^\top \bbu = \bbV^\top \bbH(\bbS) \bbx = \sum_{k=0}^K b_k \bbLambda^k \big(\bbV^\top \bbx\big) = \bbH(\bbLambda) \hat{\bbx}.
\end{equation} 
The filter operation between $\hat{\bbx}$ and $\hat{\bbu}$ is pointwise since $\bbH(\bbLambda)$ is a diagonal matrix whose $i$th diagonal entry $h(\lambda_i) = \sum_{k=0}^K b_k \lambda_i^k$ represents the filter response evaluated at $[\hat{\bbx}]_i$ in the frequency domain, i.e., $[\hat{\bbu}]_i = h(\lambda_i) [\hat{\bbx}]_i$ for $i =1,\ldots,N$. This motivates to define the \emph{filter frequency response} as an analytic function $h(\lambda) = \sum_{k=0}^K h_k \lambda^k$ on a graph frequency variable $\lambda$. The shape of $h(\lambda)$ is entirely determined by filter parameters $\ccalB$, and a specific graph $(\ccalG, \bbS)$ only instantiates specific eigenvalues $\{ \lambda_i \}_{i=1}^N$ on $\lambda$. We proceed to introduce the \emph{integral Lipschitz filter} and the \emph{Lipschitz nonlinearity}.
\begin{definition}\label{def:LipschitzFilter}
Consider the filter frequency response $h(\lambda)$ satisfying $|h(\lambda)|\le 1$. The filter is integral Lipschitz if for any frequencies $\lambda_1, \lambda_2 \in \mathbb{R}$, there exists a constant $C_L>0$ such that \vspace{-2mm}
\begin{equation}\label{eq:LipschitzFilter}
\left| h(\lambda_1)-h(\lambda_2) \right| \leq C_L \frac{|\lambda_{2}-\lambda_{1}|}{|\lambda_{1}+\lambda_{2}|/2}.
\end{equation}
\end{definition}\vspace{-2mm}
\begin{definition}\label{def:LipschitzNonlinearity}
The nonlinearity $\sigma(\cdot)$ satisfying $\sigma(0)=0$ is Lipschitz if for any $a, b \in \mathbb{R}$, there exists a constant $C_\sigma > 0$ such that\vspace{-1mm}
 \begin{equation}\label{eq:LipschitzNonlinear}
|\sigma(a) - \sigma(b)| \le C_\sigma|a - b|.
\end{equation}
\end{definition} \vspace{-1mm}
The integral Lipschitz filter is the one whose frequency response is Lipschitz in any interval $(\lambda_1, \lambda_2)$ with the Lipschitz constant inversely proportional to the interval midpoint $(\lambda_1 + \lambda_2)/2$. It is equivalent to require the derivative of the frequency response satisfying $|\lambda h'(\lambda)| \leq C_L$ for all $\lambda \in \mathbb{R}$. Such a condition is reminiscent of the scale invariance of wavelet transforms \cite{Daubechies92-Wavelets}, and common examples include graph wavelets in \cite{Hammond11-Wavelets, Shuman15-Wavelets}. This condition can also be enforced by means of penalties during training \cite{Gama20-ICASSP}. A Lipschitz nonlinearity is commonly used in neural networks, e.g., the absolute value, the ReLU, the Tanh, ect. For GCNNs with integral Lipschitz filters and Lipschitz nonlinearities, 
the following theorem characterizes the increased data diversity introduced by the graph randomness.

\begin{theorem}\label{theorem:dataDiversity}
Consider the GCNN $\bbPhi(\bbX; \bbS, \ccalA)$ consisting of integral Lipschitz filters with constant $C_L$ and Lipschitz nonlinearities with constant $C_\sigma$. Let $\bbS$ be the underlying graph and $\bbS'$ be an edge-dropped subgraph. For the graph signal $\bbX$, consider another graph signal $\bbX'$ satisfying \vspace{-1mm} 
 \begin{equation}\label{eq:anotherData}
	\bbPhi(\bbX; \bbS', \ccalA) = \bbPhi(\bbX'; \bbS, \ccalA),
\end{equation}
and measure the graph difference between $\bbS$ and $\bbS'$ w.r.t. the underlying graph $\bbS$ by considering the relative error matrix model \vspace{-2mm}
\begin{align}\label{eq:relativeError}
\bbE \in \mathbb{R}^{N \times N} : \bbS - \bbS' = \bbE \bbS + \bbS \bbE \ , \ \bbE \!=\! \bbE^{\Tr}.
\end{align}
Then, it holds that \vspace{-2mm}
 \begin{equation}\label{eq:dataDiversity}
\| \bbPhi(\bbX'; \bbS, \ccalA) - \bbPhi(\bbX; \bbS, \ccalA) \|_2 \le C \| \bbE \|_2 \| \bbX \|_2 + \ccalO(\| \bbE \|_2^2)
\end{equation}
where $C$ is a constant depending on the architecture parameters.
\end{theorem}\vspace{-1mm}

Theorem \ref{theorem:dataDiversity} states that processing the original signal $\bbX$ with the edge-dropped subgraph $\bbS'$ is equivalent to processing another signal $\bbX'$ with the underlying graph $\bbS$, and its output $\bbPhi(\bbX'; \bbS, \ccalA)$ is close to the original output $\bbPhi(\bbX; \bbS, \ccalA)$ if $\bbS'$ is similar to $\bbS$. We measure the graph similarity between $\bbS$ and $\bbS'$ with the relative error matrix model [cf. \eqref{eq:relativeError}]. Such a model 
ties the perturbation magnitude of edge dropping to the underlying graph structure by multiplying $\bbE$ with $\bbS$ where the summation is for matrix symmetry, and thus characterizes the graph similarity w.r.t. the overall topology. 
These results indicate that an edge-dropped subgraph $\bbS'$ generates an additional signal $\bbX'$, whose output is similar to that of the original signal $\bbX$ if the relative error between $\bbS'$ and $\bbS$ is small.

Given i.i.d. random edge dropping, there exist $2^{|\ccalE|}$ subgraphs with $|\ccalE|$ the number of edges. From Theorem \ref{theorem:dataDiversity}, each subgraph $\bbS_t$ sampled from the distribution $m_p(\bbS)$ yields a new signal $\bbX_{t,r}$ from the original signal $\bbX_r$. Therefore, we may reinterpret the stochastic optimization problem \eqref{eq:stochasticProblemOverS} as\vspace{-2mm}
\begin{equation} \label{eq:stochasticProblemDataAugmentation}
\begin{split}
\min_{\ccalA} \sum_{r=1}^R \sum_{t=1}^{2^{|\ccalE|}} p_t c\big(\bbPhi(\bbX_{t,r}; \bbS, \ccalA), \bby_r\big)
\end{split}
\end{equation}
where $p_t$ is the sampling probability of $\bbS_t$ from $m_p(\bbS)$. We now observe new training data $\{ \!(\bbX_{t,r},\! \bby_r)\! \}_{t,r}$ of $R \!\cdot\! 2^{|\ccalE|}$ signal-label pairs increasing the data diversity, while the augmented data $\bbX_{t,r}$ is assumed sharing the same label $\bby_r$ as the original data $\bbX_r$. This is reasonable iff $\bbX_{t,r}$ and $\bbX_r$ yield similar outputs when $\bbS_t$ and $\bbS$ capture similar topology, i.e., the relative error is mild [cf. \eqref{eq:dataDiversity}]. 

However, DropEdge samples all edges in an i.i.d. manner without accounting for the underlying graph topology, which yields a certain number of edge-dropped subgraphs substantially different from the underlying graph. In these instances, the relative errors are large, the augmented signals yield essentially different outputs, and the sharing label assumption does not hold resulting in performance degradation of data augmentation. In other words, DropEdge considers all subgraphs in $m_p(\bbS)$ with equal priority [cf. \eqref{eq:stochasticProblemOverS}] while ignoring dissimilarity between subgraphs and the underlying graph. The graph signals will be processed over some completely irrelevant subgraphs that carry little structural information inherent in graph signals during training, violating the natural reason behind the success of GCNNs. This motivates to take graph structural information into account during random edge dropping, which is exactly the proposed method.\vspace{-2mm}

\subsection{Graph connectivity} \label{subsec:graphConnectivity}

We consider graph connectivity as the key property that captures graph structure, and propose to incorporate this factor into DropEdge for more satisfactory data augmentation in which generated data does not deviate far from the original data. 
Our intuition is that if the edge-dropped subgraph $\bbS'$ maintains similar connectivity as the underlying graph $\bbS$, it preserves the overall topology, the relative error $\bbE$ [cf. \eqref{eq:relativeError}] would be small, and the augmented data that shares the same label as the original data is reasonable. To see this intuitively, note that the difference between the entry $[\bbS]_{ij}$ of the underlying graph $\bbS$ and the entry $[\bbS']_{ij}$ of the edge-dropped subgraph $\bbS'$ is given by $[\bbE\bbS+\bbS\bbE]_{ij}$. By expanding the matrix multiplication, we have\vspace{-2mm}
\begin{align} \label{eq:relativeExpansion}
	[\bbS]_{ij} - [\bbS']_{ij} \!=\! [\bbE\bbS+\bbS\bbE]_{ij} \!=\! \sum_{\tau \in \ccalN_i} [\bbE]_{i \tau} [\bbS]_{\tau j} + \sum_{\tau \in \ccalN_j} [\bbS]_{i \tau} [\bbE]_{\tau j} = \sum_{\tau \in \ccalN_i} [\bbE]_{i \tau} + \sum_{\tau \in \ccalN_j} [\bbE]_{\tau j}
\end{align} 
where $\bbS$ is assumed the adjacency matrix, and $\ccalN_i$ and $\ccalN_j$ are the sets of neighboring nodes of $n_i$ and $n_j$. It is observed that the entry difference $[\bbS]_{ij} - [\bbS']_{ij}$ is proportional to the number of neighboring nodes (edge connections) scaled by the entries of the relative error $\bbE$. As the entries of $\bbE$ grow, the entries of the underlying graph $\bbS$ and the edge-dropped subgraph $\bbS'$ become more dissimilar. However, parts of the graph with stronger connectivity would change proportionally larger than parts of the graph with weaker connectivity [cf. \eqref{eq:relativeExpansion}]. From another perspective, it is equivalent to stating that the same edge changes in parts of the graph with stronger connectivity could result from smaller relative errors than parts of the graph with weaker connectivity, and thus maintain more graph structural information. The latter emphasizes the importance of maintaining graph connectivity during random edge dropping, in order to yield small relative errors and avoid augmentation in which the generated data deviates far from the original data.\vspace{-1mm}

\section{Methodology} \label{sec:methodology}

Motivated by the above analysis, we develop TADropEdge that samples random edge-dropped subgraphs while maintaining similar topological connectivity as the underlying graph. To characterize graph connectivity, instead of considering local neighborhood of individual nodes, we take a more global perspective by considering large-scale clusters within the graph. We suppose there exist several weakly connected components in a single connected graph, referred to as \emph{clusters}---see Fig. \ref{fig:res_vals} for a motivation example with three clusters. In classification problems, for instance, $q$ is the number of node classes. In this context, inter-cluster edges are critical for establishing graph connectivity while intra-cluster edges are not. Proposition 1 in the supplementary material validates this fact by upper bounding the change in eigenvectors of the graph Laplacian induced by the change in intra-cluster edges, where the bound can be arbitrarily small as long as the edge-dropped subgraph maintains similar clustering topology. 
The main premise behind this work is to identify inter- and intra-cluster edges in the underlying graph, and to allow higher sampling probabilities for inter-cluster edges and lower sampling probabilities for intra-cluster edges. The latter not only gives the benefits as DropEdge to avoid over-fitting but also preserves overall graph topology to avoid noisy augmented data. Specifically, TADropEdge consists of three phases: 


\begin{figure*} \label{fig:clustering_example}
    \centering
    \begin{subfigure}{.18\linewidth}
    \includegraphics[scale=0.08]{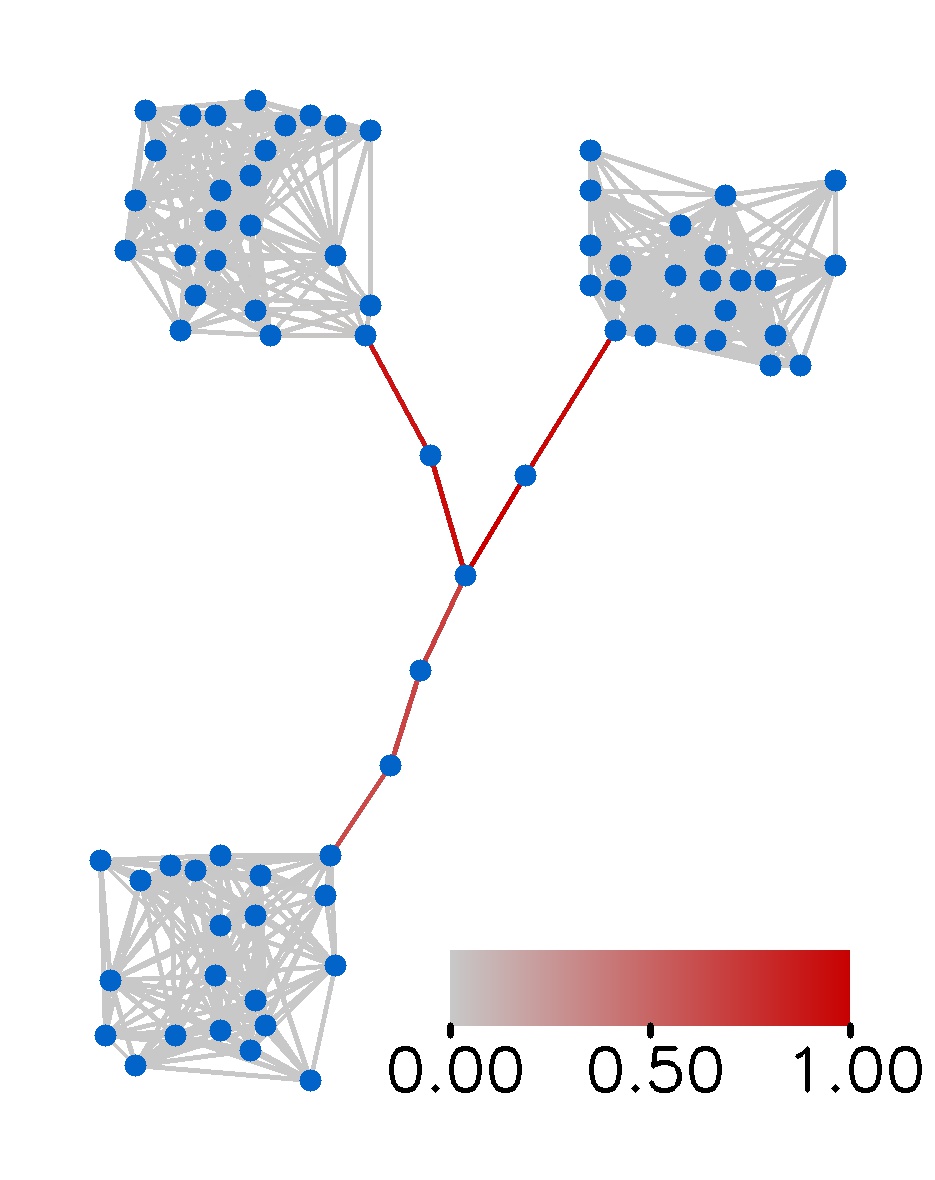}
    \caption{} \label{fig:res_vals}
    \end{subfigure}
    \begin{subfigure}{.24\linewidth}
    \includegraphics[scale=0.18]{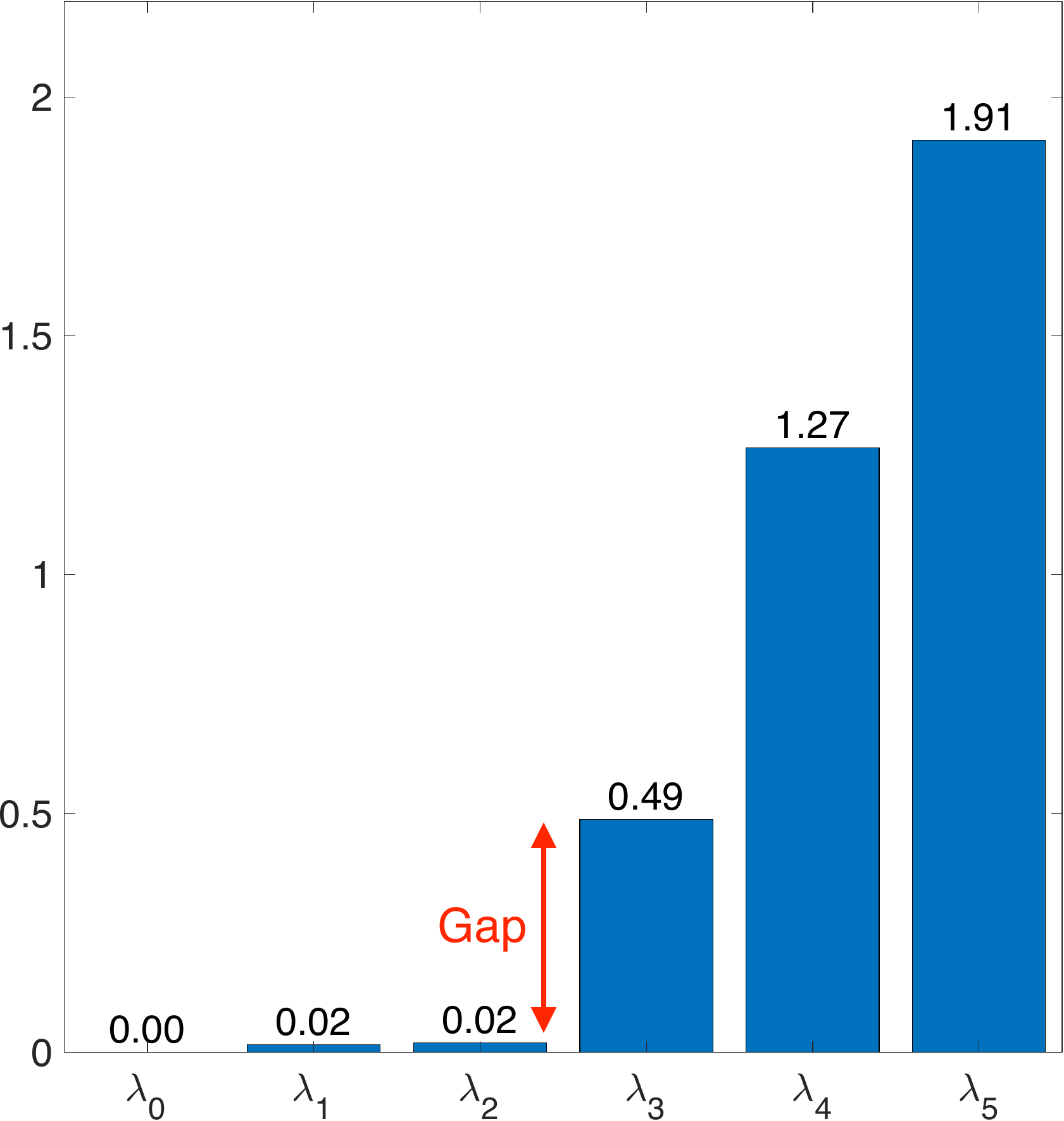}
    \caption{} \label{fig:eigvals}
    \end{subfigure}
    \begin{subfigure}{.18\linewidth}
    \includegraphics[scale=0.08]{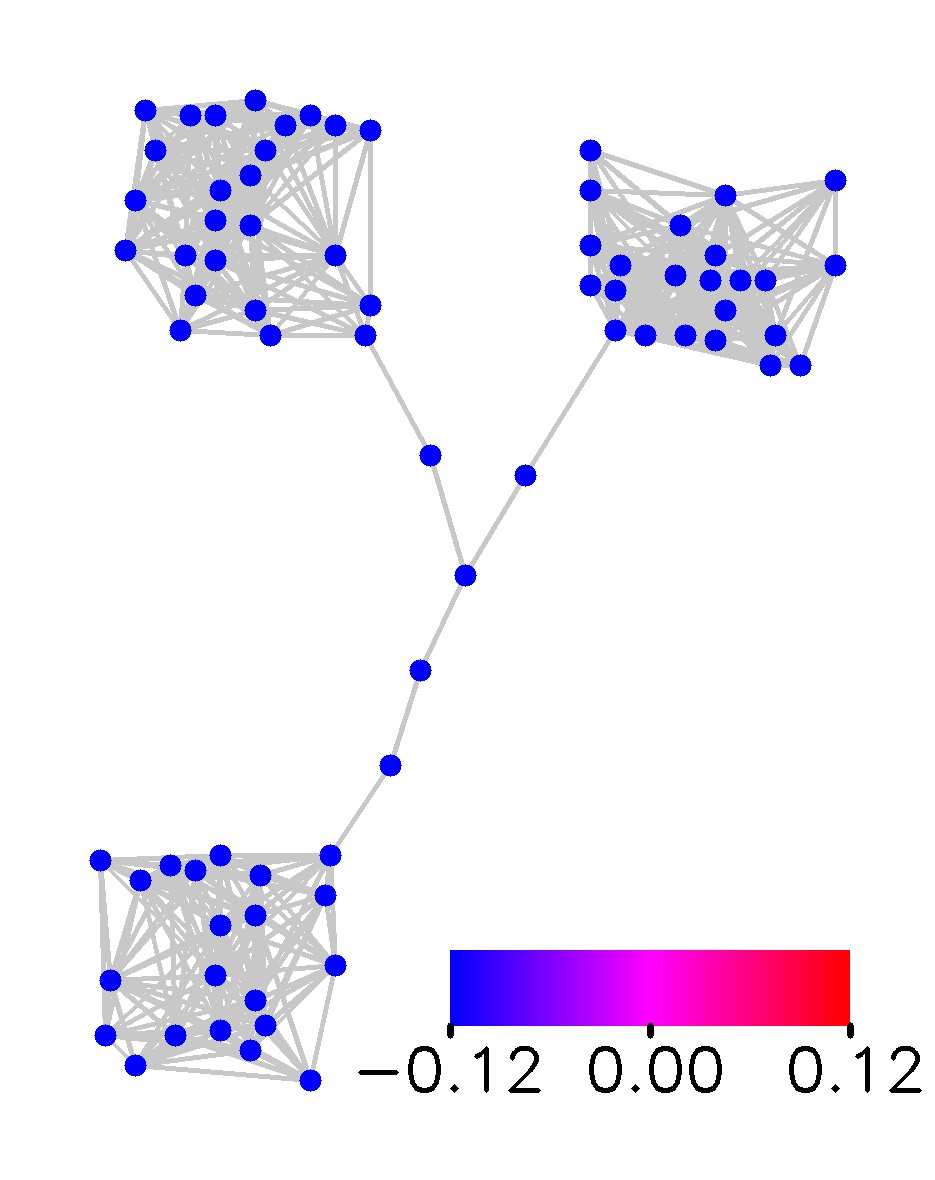}
    \caption{} \label{fig:eig_0}
    \end{subfigure}
    \begin{subfigure}{.18\linewidth}
    \includegraphics[scale=0.08]{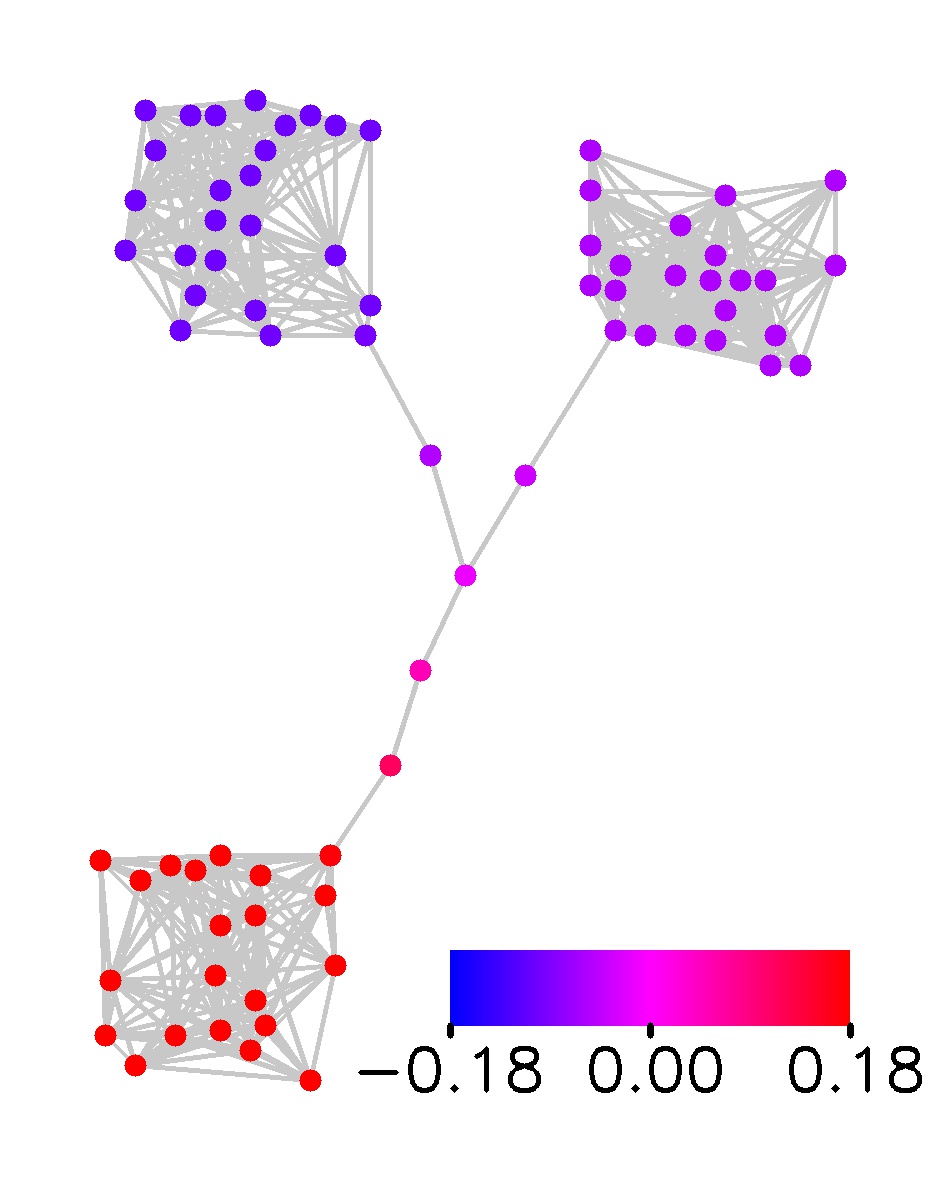}
    \caption{} \label{fig:eig_1}
    \end{subfigure}
    \begin{subfigure}{.18\linewidth}
    \includegraphics[scale=0.08]{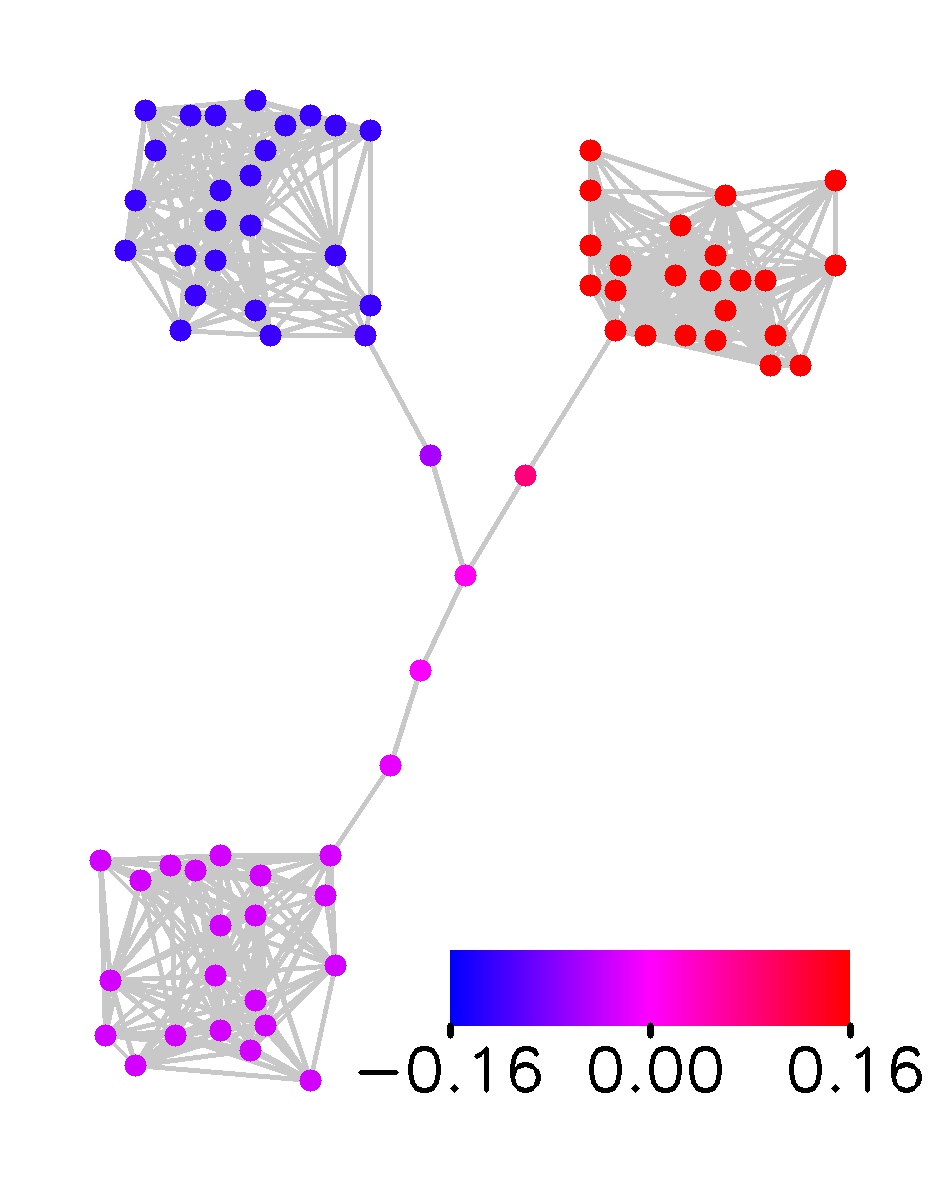}
    \caption{} \label{fig:eig_2}
    \end{subfigure}
    \caption{Fig. \ref{fig:res_vals} shows an example graph with 3 clusters. Fig. \ref{fig:eigvals} plots the first (smallest) $6$ eigenvalues. Fig. \ref{fig:eig_0}--\ref{fig:eig_2} are eigenvectors of the first 3 eigenvalues. Fig. \ref{fig:res_vals} also shows an visualization of \emph{aggregate resistance weights}.}\vspace{-6mm}
\end{figure*}


I. \textbf{Disjoint component identification}: We preprocess the underlying graph to identify \emph{disjoint components}, which are not connected to each other. In particular, there may exist several components within the underlying graph, where nodes in the same component can be connected either directly or indirectly via relay nodes while nodes between different components can never find a connecting path. We refer to the latter as disjoint components, the number of which corresponds to the number of zero eigenvalues of the graph Laplacian. We employ the disjoint-set forest to compute the union-find data structure \cite{tarjan1975efficiency}. 
The latter stores a collection of non-overlapping node sets, which allows us to determine the disjoint components and count the number of nodes in each one of them. 

II. \textbf{Edge weight computation}: We suppose there are $q$ clusters in each disjoint component (with no less than $q$ nodes). We leverage the graph spectrum to compute a positive real-valued distribution over edges, referred to as the \emph{aggregate resistance weight}, which measures the role each edge plays in being an inter- or intra-cluster edge and implies its criticality regarding to the component connectivity.

Following \cite{zhan2021barrier-journal}, we consider the first $q$ eigenvectors (correspoding to the $q$ lowest eigenvalues) of the graph Laplacian. Each eigenvector corresponds to a mode with relatively uniform value distribution over each cluster---see Fig. \ref{fig:eig_0}--\ref{fig:eig_2} for $q=3$ eigenvectors / modes. Given $q$ linearly independent modes, the uniform value of each cluster differs from the uniform value of its neighboring clusters in at least one mode. As such, the value difference between nodes that connect different clusters will be non-zero while the value difference between nodes within a single cluster will be very close to zero. This observation motivates to identify the inter- or intra-cluster edge based on the eigenvector value difference between its parent nodes across the first $q$ eigenvectors. In particular, let $\breve{\bbV} = [\mathbf{u}_{0}~\mathbf{u}_{1}~\cdots~\mathbf{u}_{q-1}] \in \mathbb{R}^{N\times q}$ be a matrix in which the columns are the first $q$ eigenvectors and $\bbD \in \mathbb{R}^{N\times M}$ be the incidence matrix that satisfies $[\bbD]_{i_m m}=1$, $[\bbD]_{j_m m}=-1$ if $e_m = (n_{i_m}, n_{j_m}) \in \ccalE$ and $[\bbD]_{im}=0$ for all $i \ne i_m, j_m$. We define aggregate resistance weight as\vspace{-2mm}
\begin{align}\label{eq:qAggregateResistance}
	\breve{\mathbf{r}} = \text{diag}\left( \bbD^\mathsf{T} \left( \breve{\bbV} \breve{\bbV}^\mathsf{T} \right) \bbD \right)
\end{align}
where the $m$th component $[\breve{\mathbf{r}}]_m$ represents the weight of the edge $e_m = (n_{i_m}, n_{j_m})$. To be more precise, expanding \eqref{eq:qAggregateResistance} yields $[\breve{\mathbf{r}}]_m \!=\! \sum_{\ell=0}^{q-1} ([\bbu_\ell]_{i_m} \!-\! [\bbu_\ell]_{j_m})^2$ for $m=1,\ldots,M$, which represents the aggregated value difference between the edge nodes $n_{i_m}$ and $n_{i_m}$ across $q$ eigenvectors. As discussed above, this vector $\breve{\mathbf{r}} \in \mathbb{R}^m$ is higher for inter-cluster edges and lower for intra-cluster edges. Fig. \ref{fig:res_vals} illustrates an example with $q=3$ clusters, 
and more examples are shown in the supplementary material. This forms the basis of TADropEdge to preserve salient topological information. That is, the edges with higher weights being the inter-cluster edges are critical in preserving graph topology and hence are sampled with higher probabilities. During the weight computation, the lowest value is logged and set as the default value for all edges of those disjoint components with less than $q$ nodes.

III. \textbf{Topology adaptive edge dropping}: We drop graph edges adaptively based on their aggregate resistance weights. Given the fact that edges with large weights are critical for establishing topological connectivity, we sample these edges with higher probabilities or maintain them as undropped. In particular, let $\ccalG$ be the underlying graph and $p$ be the default probability. TADropEdge determines the sampling matrix $\bbP_{\ccalG, p}$ by edge weights, where the entry $[\bbP]_{i_mj_m}$ is the sampling probability of the edge $e_m = (n_{i_m},n_{j_m}) \in \ccalE$. We propose three specific adaptive sampling strategies as follows: 

(i) \text{\emph{Threshold cutoff}:} We consider a threshold $\gamma$ determined by the edge weight distribution. Graph edges are sampled at the default probability $p$ if their weights are smaller than $\gamma$, otherwise they are maintained as undropped. It fixes a small number of critical edges for maintaining graph connectivity.

(ii) \text{\emph{Division normalization}:} We normalize the edge weights to $[0,1]$ with the division function, and determine the sampling matrix $\bbP_{\ccalG,p}$ by the normalized weights. Given the edge weight $\omega$ and the function parameter $\gamma$, the edge sampling probability is $~p_{\ccalG, p} = 1 - (1 - p) * \gamma/(\gamma + \omega)$.

(iii) \text{ \emph{CDF normalization}:} We normalize the edge weights to $[0,1]$ with the cumulative distribution function (CDF), and determine the sampling matrix $\bbP_{\ccalG,p}$ by the normalized weights. Given the edge weight $\omega$ and the CDF $f(\omega)$ of $\omega$, the edge sampling probability is $~p_{\ccalG, p} = p + (1-p) * f(\omega)$. 

All strategies sample graph edges of lower weights at lower probabilities close to $p$ while sampling graph edges of higher weights at higher probabilities up to $1$, where the parameter $\gamma$ is selected based on the edge weight distribution---see implementation details in the supplementary material. 
At each training epoch $t$, TADropEdge samples an edge-dropped subgraph $\ccalG_t$ with the sampling matrix $\bbP_{\ccalG, p}$. It then replaces the shift operator $\bbS$ of the underlying graph $\ccalG$ with the sparse shift operator $\bbS_t$ of the subgraph $\ccalG_t$ in the architecture [cf. \eqref{eq:layerProcessing}] for signal propagation and parameter training\footnote{We also perform the normalization technique for the edge-dropped subgraph $\bbS_t$ following the ideal of \cite{rong2019dropedge}.}. 

\begin{remark} \normalfont
	The number of clusters $q$ is a parameter for edge weight computation. On the one hand, it is related to practical applications, for instance, $q$ is the number of classes in node classification. On the other hand, $q$ can be determined from the spectrum of the graph Laplacian. If a disjoint component has $q$ clusters, there will be a \emph{gap} from the $(q-1)$th eigenvalue to the $q$th eigenvalue---see Fig. \ref{fig:eigvals} from $\lambda_2$ to $\lambda_3$. We leverage both aspects for robust edge weight computation. In particular, given the disjoint component is desired with $\alpha$ clusters, we search for a maximal gap in the neighborhood of the $(\alpha-1)$th eigenvalue and select $q$ that corresponds to the eigenvalues up until that gap. 
\end{remark}\vspace{-2mm}

\subsection{Discussions}\label{subsec:discussion}\vspace{-1mm}

\textbf{Variance reduction.} TADropEdge samples a number of critical edges with higher probabilities or keeps them undropped, which gets rid of irrelevant noisy subgraphs and thus reduces the graph randomness throughout the architecture. Moreover, it generates the edge-dropped subgraphs that maintain overall topology, which propagate graph signals similarly as the underlying graph and produce close output features. Both aspects lead to reduced output / gradient variance, which mitigates the training difficulty compared to i.i.d. DropEdge.

\textbf{Over-smoothing.} While seminal to establish more satisfactory data augmentation and improve generalization performance, TADropEdge may weaken the capacity of alleviating over-smoothing for deep GNNs. Over-smoothing is a phenomenon that output features of GNNs converge to a subspace that is only relevant to graph topology but independent to input signals, as the architecture depth increases \cite{li2018deeper, oono2019asymptotic}. Since TADropEdge exploits graph structural information during random edge dropping, the sampled subgraphs have similar overall topology though with different local structures. They will propagate graph signals in a similar way as the underlying graph, and thus output features may converge faster to the limiting subspace compared to i.i.d. DropEdge. We therefore consider TADropEdge as a trade-off between over-fitting and over-smoothing. It is worth mentioning that increasing the architecture depth typically degrades performance even if over-smoothing is alleviated with DropEdge as observed from \cite{rong2019dropedge,luo2021learning}, and TADropEdge has already achieved higher accuracies on shallow architectures with efficient implementation (Section \ref{experiments}). \vspace{-2mm}

\section{Experiments}\label{experiments}\vspace{-2mm}

We evaluate the proposed TADropEdge on node-level and graph-level classification problems with real-life and synthetic datasets, and compare it with DropEdge and existing state-of-the-art models. The experimental results show superior classification accuracies of TADropEdge corroborating theory. The supplementary material contains implementation details and more experimental results.

\subsection{Node-level classification} \vspace{-2mm}

\begin{figure*}%
\centering
\begin{subfigure}{0.25\columnwidth}
	\includegraphics[width=1.0\linewidth, height = 0.7\linewidth]{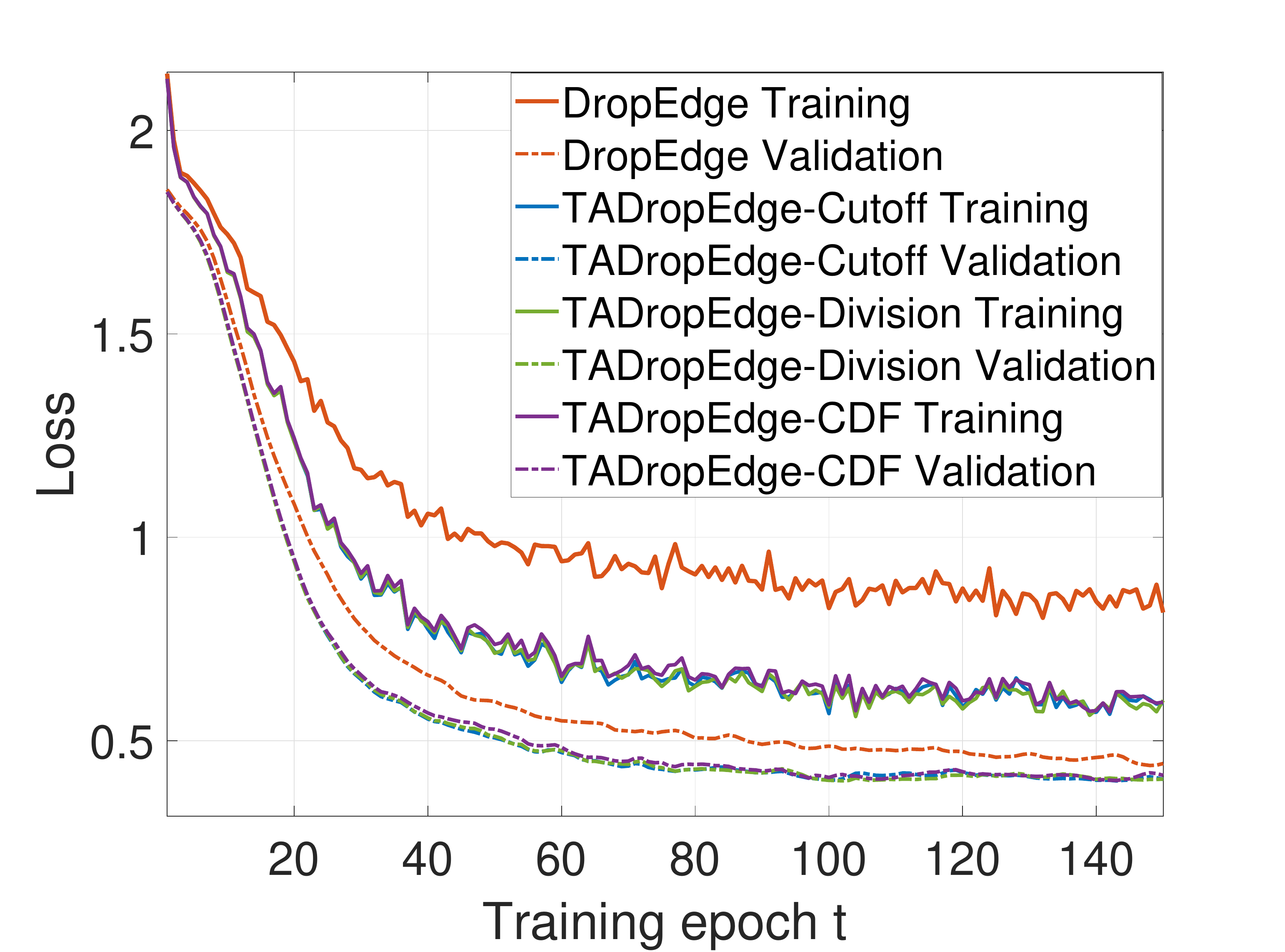}%
	\caption{}%
	\label{fig0}%
\end{subfigure}\hfill\hfill%
\begin{subfigure}{0.25\columnwidth}
\includegraphics[width=1.0\linewidth, height = 0.7\linewidth]{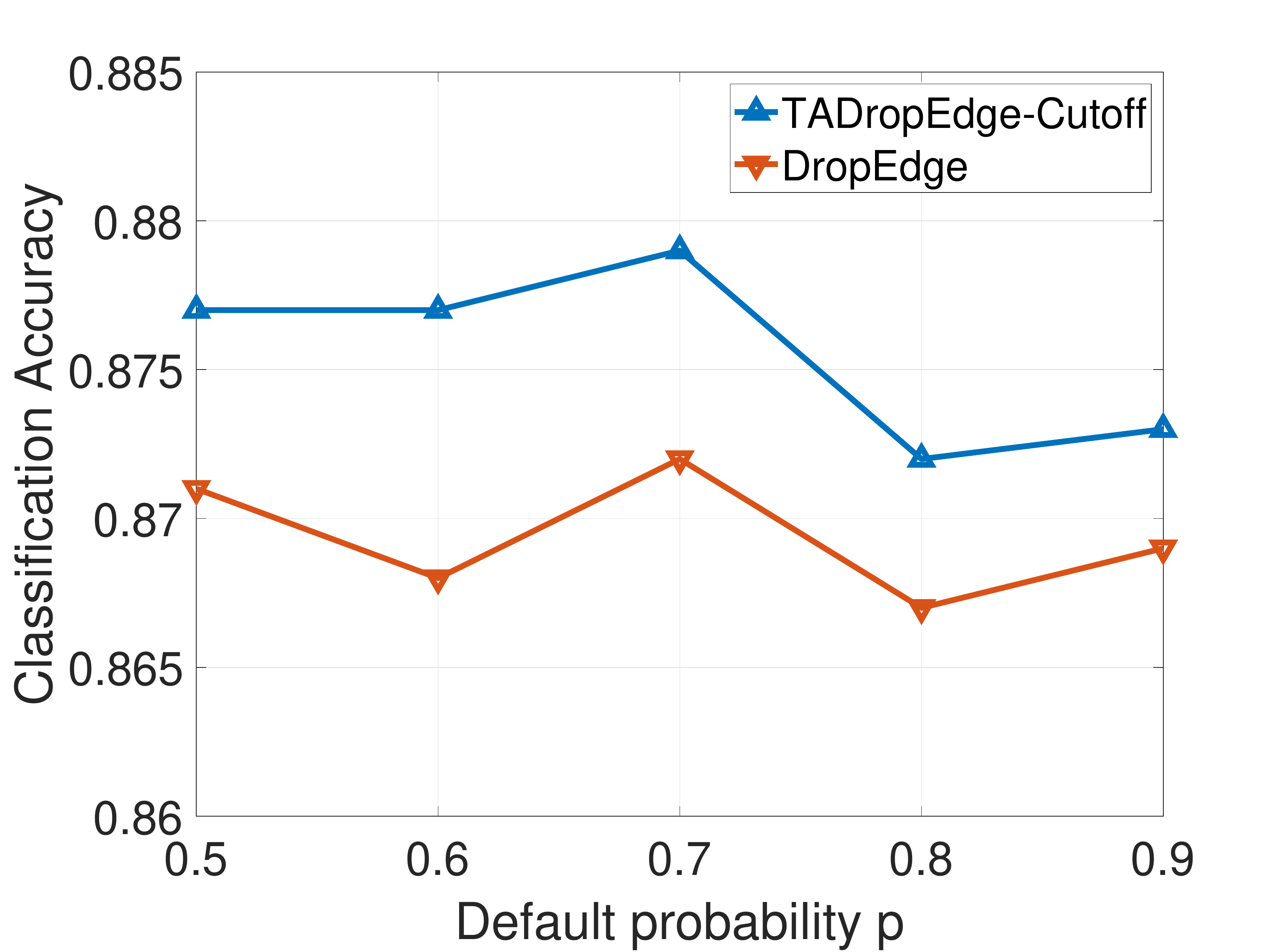}%
\caption{}%
\label{fig1}%
\end{subfigure}\hfill\hfill%
\begin{subfigure}{0.25\columnwidth}
\includegraphics[width=1.0\linewidth,height = 0.7\linewidth]{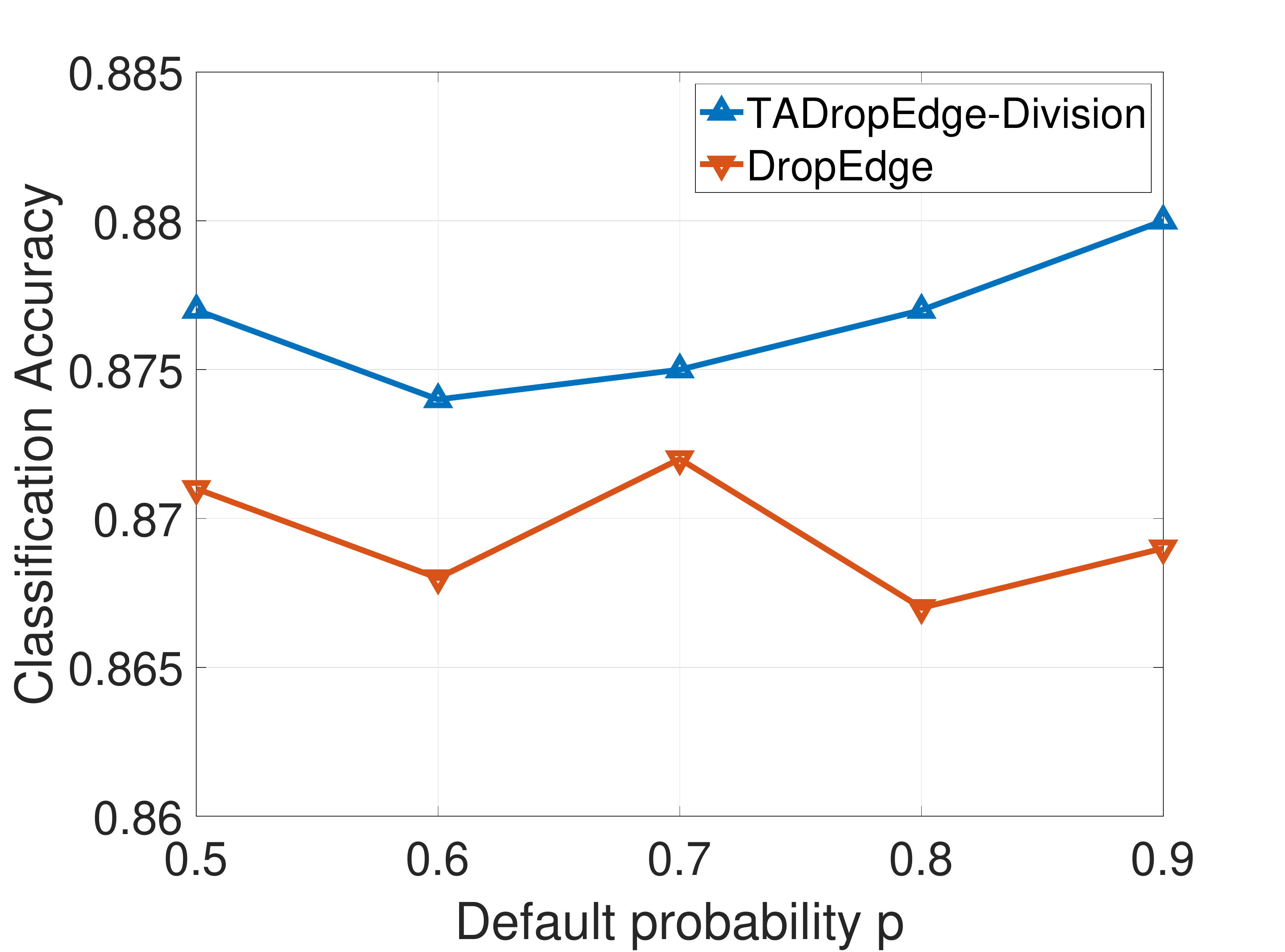}%
\caption{}%
\label{fig2}%
\end{subfigure}\hfill\hfill%
\begin{subfigure}{0.25\columnwidth}
\includegraphics[width=1.0\linewidth,height = 0.7\linewidth]{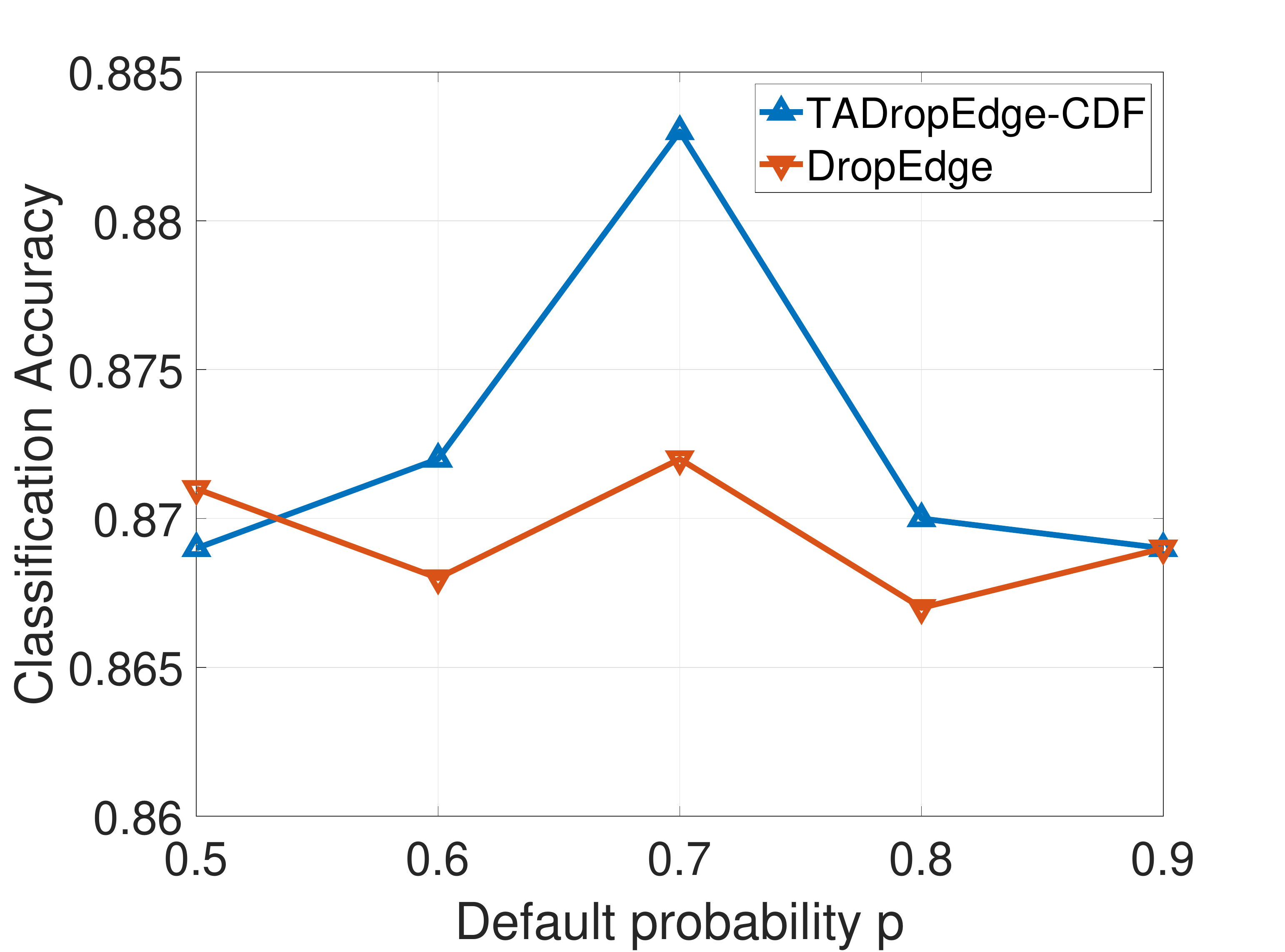}%
\caption{}%
\label{fig3}%
\end{subfigure}%
\caption{Performance comparison between TADropEdge and DropEdge for $3$-layered GCN on Cora dataset. (a) Converging behaviors. (b) Threshold cutoff. (c) Division normalization. (d) CDF normalization. }\label{fig:vary_case_study}\vspace{-6mm}
\end{figure*}

We consider the node classification in citation networks. The problem considers three citation datasets with varying graph sizes and feature types: Cora, CiteSeer and Pubmed \cite{sen2008collective}. In these datasets, graph nodes are papers and edges are citations. The node feature is a sparse bag-of-words vector, the node label is the paper topic, and the graph is undirected with a symmetric adjacency matrix. The goal is to find out the topic of a given paper. We conduct the full-supervised training following \cite{huang2018adaptive, rong2019dropedge}. 
%
\begin{table}[t]\footnotesize
	\caption{Performance comparison between TADropEdge and DropEdge} 
	\centering 
	\begin{tabular}{|l | c| p{0.5cm} p{0.5cm} p{0.6cm}| p{0.5cm} p{0.5cm} p{0.6cm}| p{0.5cm} p{0.5cm} p{0.6cm}|} 
		\hline 
		&  &  \multicolumn{3}{|c|}{GCN} & \multicolumn{3}{|c|}{JKNet} & \multicolumn{3}{|c|}{IncepGCN}  \\
		Dataset & Method & ~~2L & ~~3L & ~~4L & ~~2L & ~~3L & ~~4L & ~~2L & ~~3L & ~~4L
		\\
		\hline 
		&Original &0.868 & 0.866 & 0.860 & - & 0.869 & 0.861 & - & 0.870 & 0.862 \\
		&DropEdge &0.872 & 0.872 & 0.874 &- & 0.878 & 0.877 &- & 0.875 & 0.879 \\
		Cora & TADropEdge-Cutoff
		& \textbf{0.879} & 0.879 & 0.877& - & 0.883 & 0.883 & - & \textbf{0.889} & 0.881 \\& TADropEdge-Division & 0.878 & 0.880 & 0.878 & - & \textbf{0.886} & \textbf{0.886} & - & 0.886 & \textbf{0.884} \\& TADropEdge-CDF & \textbf{0.879} & \textbf{0.883} & \textbf{0.879} & - & \textbf{0.886} & 0.883 & - & 0.887 & 0.883 \\ \hline
		&Original &0.794 & 0.789 & 0.777 &- & 0.793 & 0.789 &- & 0.787 & 0.794 \\
		&DropEdge &0.802 & 0.798 & 0.783 &- & 0.796 & 0.791 &- & 0.794 & 0.796 \\
		CiteSeer & TADropEdge-Cutoff
		& 0.812 & 0.805 & \textbf{0.791} &- & \textbf{0.807} & 0.796 &- & 0.804 & \textbf{0.801} \\& TADropEdge-Division
		& \textbf{0.814} & 0.804 & 0.790 &- & 0.806 & 0.796 &- & \textbf{0.805} & \textbf{0.801} \\& TADropEdge-CDF
		& \textbf{0.814} & \textbf{0.807} & \textbf{0.791} &- & 0.806 & \textbf{0.797} &- & 0.803 & 0.800 \\ \hline
		&Original &0.904 & 0.894 & 0.891 &- & 0.895 & 0.890 &- & 0.896 & 0.895 \\
		&DropEdge &0.909 & 0.907 & 0.904 &- & 0.907 & 0.903 &- & 0.904 & 0.906 \\
		Pubmed & TADropEdge-Cutoff
		& 0.912 & 0.912 & 0.906 &- & \textbf{0.911} & 0.908 &- & \textbf{0.910} & \textbf{0.911} \\& TADropEdge-Division
		& 0.911 & 0.912 & \textbf{0.908} &- & 0.910 & 0.907 &- & \textbf{0.910} & \textbf{0.911} \\& TADropEdge-CDF
		& \textbf{0.914} & \textbf{0.915} & 0.905 &- & 0.910 & \textbf{0.910} &- & 0.908 & 0.909 \\
		\hline 
	\end{tabular}
	\label{tab:ComparisonDropEdge}
	\vspace{-5mm}
\end{table}

\textbf{Comparison with DropEdge.} We start by providing a case study to demonstrate how TADropEdge improves performance against i.i.d. DropEdge, where we focus on the dataset Cora and the backbone GCN of $3$ layers \cite{Kipf2017}. Fig. \ref{fig0} shows converging behaviors of compared methods at $p=0.7$. TADropEdge converges faster with lower training / validation loss. The former is due to variance reduction (Section \ref{subsec:discussion}) and the latter is because TADropEdge accounts for graph structural information. Fig. \ref{fig1}-\ref{fig3} report the classification accuracy comparison between TADropEdge and DropEdge under varying default probabilities $p$. While different adaptive sampling strategies yield (slightly) different performance, 
TADropEdge consistently outperforms DropEdge in almost all scenarios. 
These results validate the importance of maintaining graph topology during random edge dropping.

We then formally compare TADropEdge and DropEdge on all three datasets: Cora, CiteSeer and Pubmed. We consider three backbones: GCN \cite{Kipf2017}, IncepGCN \cite{rong2019dropedge} and JKNet \cite{Xu19-GIN}, where the architecture depth varies from $2$ to $4$ layers for efficient implementation\footnote{Deeper architectures increase computational complexity but degrades performance due to over-smoothing in most cases \cite{rong2019dropedge, luo2021learning}, such that we focus on shallow architectures in this paper.}. We follow the experimental setting in \cite{rong2019dropedge} to perform a random hyper-parameter search for each model---see details in the supplementary material. Table \ref{tab:ComparisonDropEdge} summaries the results. It is observed that TADropEdge improves classification accuracy consistently in all cases. While different adaptive sampling strategies affect performance slightly, all of them outperform DropEdge corroborating the effectiveness of TADropEdge on exploiting graph structural information during random edge dropping. The performance improvement is more remarkable for GNNs of lower layers, which 
can be explained by the fact that TADropEdge may weaken the ability of alleviating over-smoothing in deep GNNs, 
as discussed in Section \ref{subsec:discussion}.
\begin{table}[t]\footnotesize
\caption{Performance comparison between TADropEdge and SOTA methods} 
\centering 
\begin{tabular}{|l | c| c| c| c|} 
\hline 
  & Method & Cora & CiteSeer & Pubmed
\\ [0.5ex]
\hline 
 &FastGCN &0.8500 & 0.7760 & 0.8800 \\
 &AS-GCN &0.8744 & 0.7966 & 0.9060 \\
SOTA methods &GraphSAGE
& 0.8220 & 0.7140 & 0.8710 \\& NeuralSparse
& 0.8230 & 0.7730 & 0.8500 \\& PTDNeT
& 0.8250 & 0.7840 & 0.8540 \\ \hline
 & TADropEdge-GCN & 0.8830 (3L) & \textbf{0.8140 (2L)} & \textbf{0.9150 (3L)} \\
TADropEdge & TADropEdge-JKNet
& 0.8860 (3L / 4L) & 0.8070 (3L) & 0.9110 (3L) \\& TADropEdge-IncepGCN
& \textbf{0.8890 (3L)} & 0.8050 (3L) & 0.9110 (4L) \\ \hline
\end{tabular}
\label{tab:ComparisonSOTAs}
\vspace{-5mm}
\end{table}

\textbf{Comparison with SOTAs.} We select the best performance of each backbone with TADropEdge, and compare with existing state-of-the-art (SOTA) models. The latter include sampling methods: FastGCN \cite{chen2018fastgcn}, AS-GCN \cite{huang2018adaptive}, GraphSAGE \cite{hamilton2017inductive} and supervised sparsification methods: NeuralSparse \cite{zheng2020robust} and PTDNET \cite{luo2021learning}. 
For sampling methods, we reuse the results reported in \cite{huang2018adaptive}; For sparsification methods, we utilize the Tensorflow implementation in \cite{luo2021learning}. Note that we follow the full-supervised setting in\cite{huang2018adaptive, rong2019dropedge}, which is different from that in \cite{luo2021learning}. We summarize the comparison results in Table \ref{tab:ComparisonSOTAs}. TADropEdge exhibits the best performance against SOTA methods on all datasets. 
The performance improvements on Cora and Citeseer are more significant than that on Pubmed. This is because the underlying graphs of Cora and Citeseer have better clustering properties than that of Pubmed and thus, graph structural information is more important during random edge dropping. For most cases, the best accuracy of TADropEdge is achieved under the depth of $2$ or $3$ layers. This indicates that TADropEdge is able to achieve superior performance with efficient implementation.\vspace{-4mm}
%

\begin{table}[h]\footnotesize
	\begin{center}
		\caption{Performance comparison on source localization between TADropEdge and DropEdge.}
		\label{table1}
		\begin{tabular}{|l|c|c|c|c|c|}
			\hline
			Method & Classification accuracy \\ \hline
			GCNN & $0.820$ \\ \hline
			GCNN with DropEdge & $0.836$ \\ \hline
			GCNN with TADropEdge-Cutoff &  $\textbf{0.860}$ \\ \hline
			GCNN with TADropEdge-Division &  $0.856$ \\ \hline
			GCNN with TADropEdge-CDF &  $0.848$ \\ \hline
		\end{tabular}
	\label{tab:ComparisonSource}
	\end{center}  \vspace{-6mm}
\end{table}

\subsection{Graph-level classification}\vspace{-2mm}

We consider the graph-level classification as source localization. The problem considers the signal diffusion process over the stochastic block model (SBM) graph of $N$ nodes equally divided into $C$ communities, where the inter- and intra-community link probabilities are $0.2$ and $0.6$ respectively. The initial source signal is a Kronecker delta $\bbdelta_s = [\delta_1, \ldots, \delta_N]^\top \in \{ 0,1 \}^N$ with $\delta_s \neq 0$ at the source node $s \in \{ s_1, \ldots, s_C \}$. The diffused signal at time $t$ is $\bbx_{st} = \bbS^t \bbdelta_s + \bbn$ with $\bbS$ the normalized adjacency matrix and $\bbn$ the normal noise. The goal is to determine the source community of a given diffused signal. We consider a single-layered GCNN \cite{Fernando2019}, and compare TADropEdge with DropEdge and the original model. We measure performance as the classification accuracy, and follow \cite{rong2019dropedge} to perform a random hyper-parameter search to report the best accuracy for each method---see details in the supplementary material.
Table \ref{tab:ComparisonSource} shows the results. We see that both TADropEdge and DropEdge outperform the original model, which implies random edge dropping successfully augments the training data. TADropEdge exhibits better performance than DropEdge, where the threshold cutoff obtains the highest classification accuracy. As previously explained, this is because TADropEdge maintains the overall topology of the underlying graph during random edge dropping and avoids noisy augmented data. The performance improvement of TADropEdge is more significant compared to the node classification in citation networks. We attribute this behavior to the fact that the SBM graph has better clustering properties, and its topology is more vulnerable to i.i.d. edge dropping.\vspace{-2mm}

\section{Conclusion}\label{C}\vspace{-2mm}

This paper proposed Topology Adaptive Edge Dropping method to improve generalization performance of graph neural networks. TADropEdge accounts for the underlying graph topology during random edge dropping, such that the edge-dropped subgraphs maintain the structural information inherent in graph signals and avoid noisy data augmentation for performance enhancement. TADropEdge consists of three phases: 1) Identify disjoint components within the underlying graph; 2) Assign connectivity-relevant weights to edges for each disjoint component; 3) Normalize the edge weights and drop graph edges adaptively based on the normalized weights. The three phases identify critical edges that establish the graph connectivity, and sample them with higher probabilities to maintain the overall topology of the underlying graph. Considerable experiments on real-life and synthetic datasets validate that TADropEdge consistently promotes performance and learns robust GNNs.


\bibliographystyle{IEEEtran}
\bibliography{myIEEEabrv,biblioDropEdge}

\newpage

\appendix

\begin{table*}[!]
	\centering
	{\Large Supplementary Materials for}:~
	{\Large  Training Robust Graph Neural Networks with Topology Adaptive Edge Dropping}
\end{table*}

\section{Proof of Theorem 1}

We need the following lemma in the proof.

\begin{lemma} \label{lemma1}
	Consider the integral Lipschitz filter $\bbH(\bbS)$ [Def. 1 in the full paper] with one-dimensional input signal $\bbx \in \mathbb{R}^{N \times 1}$, one-dimensional output feature $\bbH(\bbS)\bbx \in \mathbb{R}^{N \times 1}$ and integral Lipschitz constant $C_L$. Let $\bbS = \bbV \bbLambda \bbV^\top$ be the underlying graph of $N$ nodes, $\bbS'$ be an edge-dropped subgraph, and $\bbE= \bbU \bf{\Theta}\bbU^\top$ be the relative error matrix between $\bbS$ and $\bbS'$ [cf. (8) in the full paper]. 
	Then, for any graph signal $\bbx$, the output difference between filters $\bbH(\bbS)$ and $\bbH(\bbS')$ satisfies
	\begin{equation} \label{eq:lemma1}
		\| \bbH(\bbS)\bbx - \bbH(\bbS')\bbx\|_2 \le 2C_L(1 + \sigma \sqrt{N}) \|\bbE\|_2 \| \bbx \|_2 + \ccalO(\|\bbE\|_2^2)
	\end{equation}
	where $\delta = (\| \bbU - \bbV \|_2+1)^2-1$ implies the eigenvector misalignment between $\bbS$ and $\bbE$.
\end{lemma} 
\begin{proof}
	Given the one-dimensional input signal $\bbx \in \mathbb{R}^{N \times 1}$ and the one-dimensional output feature $\bbH(\bbS)\bbx \in \mathbb{R}^{N \times 1}$, the graph filter output can be represented as
	\begin{align}\label{prooflemma1:eq1}
		\bbH(\bbS)\bbx = \sum_{k=0}^K b_k \bbS^k \bbx
	\end{align}
	where $\ccalB = \{ b_0, \ldots, b_K \}$ are filter parameters. The output difference between filters $\bbH(\bbS)$ and $\bbH(\bbS')$ is then given by
	\begin{align}\label{prooflemma1:eq2}
		\big\|\bbH(\bbS)\bbx - \bbH(\bbS')\bbx\big\|_2 = \big\|\sum_{k=0}^K b_k \bbS^k \bbx - \sum_{k=0}^K b_k {\bbS'}^k \bbx \big\|_2
	\end{align}
	We then refer to Theorem 2 in \cite{gama2020stability} to complete the proof.
\end{proof}

\begin{proof}[Proof of Theorem 1]
	Since $\bbPhi(\bbX;\bbS', \ccalA) = \Phi(\bbX';\bbS,\ccalA)$ [cf. (7) in the full paper], we have
	\begin{equation}\label{proofthm2:eq0} 
		\| \Phi(\bbX';\bbS,\ccalA) - \Phi(\bbX;\bbS,\ccalA) \|_2 = \| \Phi(\bbX;\bbS',\ccalA) - \bbPhi(\bbX;\bbS, \ccalA) \|_2.
	\end{equation}
	Consider the layered architecture of the GCNN, where the input $\bbX \in \mathbb{R}^{N \times F}$ is an $F$-dimensional graph signal and the output $\bbPhi(\bbX;\bbS, \ccalA) \in \mathbb{R}^{N \times F_L}$ is an $F_L$-dimensional graph signal. We consider the norms of $\bbX$ and $\bbPhi(\bbX;\bbS, \ccalA)$ as
	\begin{align}\label{proofthm2:eq05} 
		\|\bbX\|_2 = \sum_{f=1}^F \| [\bbX]_f \|_2~~~\text{and}~~~\|\bbPhi(\bbX;\bbS, \ccalA)\|_2 = \sum_{f=1}^{F_L}\| [\bbPhi(\bbX;\bbS, \ccalA)]_f \|_2
	\end{align}
	where $[\cdot]_f$ represents the $f$th column. At each layer $\ell$, the graph convolution [cf. (1) in the full paper] is equivalent to applying $F_{\ell-1}F_\ell$ one-dimensional input and one-dimensional output filters [cf. \eqref{prooflemma1:eq1}], i.e., we have
	\begin{equation}\label{proofthm2:eq1} 
		\left[\sum_{k=0}^{K} \bbS^{k} \bbX_{\ell-1} \bbB_{\ell, k}\right]_f = \sum_{g=1}^{F_{\ell-1}} \sum_{k=0}^K  b_{\ell, k}^{fg} \bbS^k [\bbX_{\ell-1}]_{g},~\for~f=1,\ldots,F_\ell 
	\end{equation}
	where $b_{\ell, k}^{fg} = [\bbB_{\ell, k}]_{fg}$ is the $(f,g)$th entry of the matrix $\bbB_{\ell, k} \in \mathbb{R}^{F_\ell \times F_{\ell-1}}$. We denote by $\bbH_\ell^{fg}(\bbS)[\bbX_{\ell-1}]_{g}= \sum_{k=0}^K b_{\ell,k}^{fg} \bbS^k [\bbX_{\ell-1}]_{g}$ and $\bbx^g_{\ell-1} = [\bbX_{\ell-1}]_{g}$ as concise notations. By substituting \eqref{proofthm2:eq1} into the architecture [cf. (2) in the full paper], the output difference is
	\begin{align} \label{eqn:thm13}
		\|\bbPhi(\bbX;\! \bbS,\! \ccalA)\!-\!\bbPhi(\bbX;\! \bbS',\! \ccalA)\|_2 &= \sum_{f=1}^{F_L} \big\| \sigma\big( \sum_{g=1}^{F_{L-1}} \bbH_L^{fg}(\bbS)\bbx^g_{L-1} \big) - \sigma\big( \sum_{g=1}^{F_{L-1}} \bbH_L^{fg}(\bbS'){\bbx'}^g_{L-1} \big) \big\|_2 \nonumber\\
		&\le \sum_{f=1}^{F_L} C_\sigma \sum_{g=1}^{F_{L-1}}  \big\| \bbH_L^{fg}(\bbS)\bbx^g_{L-1} - \bbH_L^{fg}(\bbS'){\bbx'}^g_{L-1} \big\|_2
	\end{align}
	where $\cdot'$ represents an operation acting on $\bbS'$ instead of $\bbS$, and the second inequality is because of the Lipschitz nonlinearity and the triangle inequality. By adding and substracting $\bbH_L^{fg}(\bbS')\bbx^g_{L-1}$ into the terms inside the norm, we have
	\begin{align} \label{eqn:thm14}
		\big\| \bbH_L^{fg}\!(\bbS)\bbx^g_{L\!-\!1} \!\!\!-\! \bbH_L^{fg}\!(\bbS'){\bbx'}^g_{L\!-\!1} \!\big\|_2\! \!&\le\! \big\| \bbH_L^{fg}\!(\bbS)\bbx^g_{L\!-\!1} \!\!\!-\! \bbH_L^{fg}\!(\bbS')\bbx^g_{L\!-\!1} \!\big\|_2 \!\!\!+\!\! \big\|\bbH_L^{fg}\!(\bbS')\bbx^g_{L\!-\!1} \!\!\!-\! \bbH_L^{fg}\!(\bbS'){\bbx'}^g_{L\!-\!1} \!\big\|_2 \nonumber \\
		& \le \!\big\| \bbH_L^{fg}\!(\bbS)\bbx^g_{L\!-\!1} \!\!-\! \bbH_L^{fg}\!(\bbS')\bbx^g_{L\!-\!1}\! \big\|_2 \!\! +\! \| \bbH_L^{fg}\!(\bbS') \| \big\|\bbx^g_{L\!-\!1} \!\!-\! {\bbx'}^g_{L\!-\!1}\! \big\|_2.
	\end{align}
	For the first term in \eqref{eqn:thm14}, by using Lemma \ref{lemma1}, we get
	\begin{equation} \label{eqn:thm15}
		\big\| \bbH_L^{fg}\!(\bbS)\bbx^g_{L\!-\!1} \!-\! \bbH_L^{fg}\!(\bbS')\bbx^g_{L\!-\!1} \big\|_2 \! \le  2C_L(1 + \sigma \sqrt{N}) \|\bbx^g_{L\!-\!1}\|_2 \| \bbE \|_2 + \ccalO(\| \bbE \|_2^2).
	\end{equation}
	For the norm of the $L$th layer input $\|\bbx^g_{L\!-\!1}\|_2$, we observe that 
	\begin{equation} \label{eqn:thm16}
		\|\bbx^g_{L\!-\!1}\|_2 = \big\| \sigma\big( \sum_{u=1}^{F_{L-2}} \bbH_{L-1}^{gu}(\bbS)\bbx^u_{L-2} \big) \big\|_2 \!\le\!  C_\sigma \! \sum_{u=1}^{F_{L-2}} \big\| \bbH_{L-1}^{gu}(\bbS)\bbx^u_{L-2} \big\|_2 \!\le\! C_\sigma \! \sum_{u=1}^{F_{L-2}} \big\| \bbx^u_{L-2} \big\|_2
	\end{equation}
	where we use the triangle inequality, followed by the bound on filters \cite{Ortega18-GSP}, i.e., the filter frequency response $|h^{gu}_{L-1}(\lambda)| = \big| \sum_{k=0}^K h_{L-1,k}^{gu} \lambda^k \big| \le 1$. Following this recursion yields
	\begin{equation} \label{eqn:thm17}
		\|\bbx^g_{L\!-\!1}\|_2 \le C_\sigma^{L-1} \prod_{\ell = 1}^{L-2} F_\ell \sum_{u=1}^F \big\| \bbx^u_0 \big\|_2 = C_\sigma^{L-1} \prod_{\ell = 1}^{L-2} F_\ell \| \bbX \|_2
	\end{equation}
	with $\| \bbx^u_0 \|_2 = \| [\bbX]_u \|_2$ by definition and $\| \bbX \|_2 = \sum_{u=1}^F \| [\bbX]_u \|_2$ from \eqref{proofthm2:eq05}. By substituting \eqref{eqn:thm17} into \eqref{eqn:thm15}, we have
	\begin{equation} \label{eqn:thm18}
		\big\| \bbH_L^{fg}\!(\bbS)\bbx^g_{L\!-\!1} \!-\! \bbH_L^{fg}\!(\bbS')\bbx^g_{L\!-\!1} \big\|_2 \le  2C_L(1 + \sigma \sqrt{N}) C_\sigma^{L-1}\! \prod_{\ell = 1}^{L-2} F_\ell \| \bbX \|_2 \| \bbE \|_2 + \ccalO(\| \bbE \|_2^2).
	\end{equation}
	For the second term in \eqref{eqn:thm14}, by again using the filter bound \cite{Ortega18-GSP}, we get
	\begin{equation} \label{eqn:thm19}
		\| \bbH_L^{fg}\!(\bbS') \|_2 \big\|\bbx^g_{L\!-\!1} \!-\! {\bbx'}^g_{L\!-\!1} \big\|_2 \le \big\|\bbx^g_{L\!-\!1} \!-\! {\bbx'}^g_{L\!-\!1} \big\|_2.
	\end{equation}
	By substituting \eqref{eqn:thm18} and \eqref{eqn:thm19} into \eqref{eqn:thm14} and the latter into \eqref{eqn:thm13}, we have
	\begin{align} \label{eqn:thm110}
		&\|\bbPhi(\bbX;\! \bbS,\! \ccalA)\!-\!\bbPhi(\bbX;\! \bbS',\! \ccalA)\|_2 \\
		&\le\! \sum_{f=1}^{F_{L}}C_\sigma \sum_{g=1}^{F_{L-1}}\big\|\bbx^g_{L\!-\!1} \!-\!{\bbx'}^g_{L\!-\!1} \big\|_2 + 2C_L(1 \!+\! \sigma \sqrt{N}) C_\sigma^{L}\! \prod_{\ell = 1}^{L-1} F_\ell \| \bbX \|_2 \| \bbE \|_2 + \ccalO^2(\| \bbE \|_2). \nonumber
	\end{align}
	From \eqref{eqn:thm110}, we observe that the output difference of the $L$th layer depends on that of the $(L-1)$th layer. Repeating this recursion until the input layer and substituting the result into \eqref{proofthm2:eq0} , we complete the proof
	\begin{align} \label{eqn:thm111}
		\|\bbPhi(\bbX';\! \bbS,\! \ccalA)\!-\!\bbPhi(\bbX;\! \bbS,\! \ccalA)\| \le 2C_L(1 + \sigma \sqrt{N}) L C_\sigma^L \prod_{\ell = 1}^{L} F_\ell \| \bbX \|_2 \| \bbE \|_2 + \ccalO^2(\| \bbE \|_2)
	\end{align}
	where we use the initial condition that $\| \bbx_0^u - {\bbx'}_0^u \|_2 = \| [\bbX]_u - [\bbX]_u \|_2=0$ for all $u=1,\ldots,F$.
	
\end{proof}

\section{Clustering Property of Graph}

We characterize graph connectivity in terms of clustering within the graph, i.e., we suppose there exist several strongly connected clusters with weak inter-cluster connections. From this perspective, inter-cluster edges are more critical for establishing graph connectivity while intra-cluster edges are not. In this section, we validate this fact by providing detailed theoretical analysis.  

In particular, let $\ccalG$ be a single connected graph consisting of $q$ clusters $\{ \ccalG_\kappa\}_{\kappa=0}^{q-1}$. Each cluster $\ccalG_\kappa$ is a subgraph with the node set $\ccalV(\ccalG_\kappa) \in \ccalV$ and the edge set $\ccalE(\ccalG_\kappa) \in \ccalE$, such that $\cup_{\kappa=0}^{q-1} \ccalV(\ccalG_\kappa) = \ccalV$ and $e_m = (i_m, j_m) \in \ccalE(\ccalG_\kappa)$ if $e_m \in \ccalE$ and $n_{i_m}, n_{j_m} \in \ccalV(\ccalG_\kappa)$. We define the constituted graph 
\begin{align}\label{eq:constitutedGraph}
	\overline{\ccalG} = \cup_{\kappa=0}^{q-1}\ccalG_\kappa
\end{align}
with the same node set $\ccalV$ as $\ccalG$ and the edge set $\ccalE(\overline{\ccalG}) = \cup_{\kappa=0}^{q-1}\ccalE(\ccalG_\kappa) \in \ccalE$. The latter has $q$ disjoint clusters / subgraphs $\{ \ccalG_\kappa\}_{\kappa=0}^{q-1}$ that are not connected to each other, which can be considered as the extreme clustering property. We proceed to introduce some relevant definitions by \cite{zhan2021barrier-journal} with respect to the clustering property.

\begin{definition}[Relative Subgraph Degree]\label{def:relativeSubgraphDegree}
	Consider the underlying graph $\ccalG$ with nodes $\ccalV$ and edges $\ccalE$, and its subgraph $\ccalG_\kappa$ with nodes $\ccalV(\ccalG_\kappa)$. Let $\ccalE(\ccalG, \ccalG_\kappa) \subseteq \ccalE$ be the set of edges that connect a node inside the subgraph $\ccalV(\ccalG_\kappa)$ and a node outside the subgraph $\ccalV \backslash \ccalV(\ccalG_\kappa)$. The relative subgraph degree of $\ccalG_\kappa$ in $\ccalG$ is defined as
	\begin{align}\label{eq:relativeSubgraphDegree}
		\beta_\ccalG(\ccalG_\kappa) = \frac{1}{\sqrt{|\ccalV(\ccalG_\kappa)|}} |\ccalE(\ccalG, \ccalG_\kappa)|
	\end{align}
	where $|\ccalV(\ccalG_\kappa)|$ is the number of nodes in $\ccalV(\ccalG_\kappa)$ and $|\ccalE(\ccalG, \ccalG_\kappa)|$ is the number of edges in $\ccalE(\ccalG, \ccalG_\kappa)$.
\end{definition}

\begin{definition}[Average Relative Subgraph Degree]\label{def:averageRelativeSubgraph}
	Consider the underlying graph $\ccalG$ consisting of $q$ clusters / subgraphs $\{ \ccalG_\kappa\}_{\kappa=0}^{q-1}$, the average relative subgraph degree is defined as the root mean square of the relative subgraph degrees of the clusters [Def. \ref{def:relativeSubgraphDegree}]
	\begin{align}\label{eq:averageRelativeSubgraphDegree}
		\bar{\beta}_{\ccalG} (\ccalG_0,\ldots,\ccalG_{q-1}) = \sqrt{\frac{1}{q}\sum_{\kappa=0}^{q-1}{\big(\beta_{\ccalG}(\ccalG_\kappa)\big)^2}}
	\end{align}
	where $\gamma_{\ccalG}(\ccalG_\kappa)$ is the relative subgraph degree of the cluster $\ccalG_\kappa$ [cf. \eqref{eq:relativeSubgraphDegree}].
\end{definition}

\begin{definition}[$\alpha$-realizable $q$-partition]\label{def:alphaRelaization}
	A graph $\ccalG$ is said to have an $\alpha$-realizable $q$-partition if there exist $q$ clusters $\{\ccalG_0,\ldots,\ccalG_{q-1}\}$ such that
	\begin{align}\label{eq:alphaRealizable}
		\bar{\beta}_\ccalG(\ccalG_0,\ldots,\ccalG_{q-1}) \le \frac{\alpha \lambda_q}{\sqrt{2q}}
	\end{align}
	where $\bar{\beta}_\ccalG(\ccalG_0,\ldots,\ccalG_{q-1})$ is the average relative subgraph degree of $\ccalG$ [cf. \eqref{eq:averageRelativeSubgraphDegree}] and $\lambda_q$ is the $(q+1)$th eigenvalue (in order of magnitude) of the graph Laplacian $\bbL$ of $\ccalG$.
\end{definition}

The relative subgraph degree $\beta_\ccalG(\ccalG_\kappa)$ quantifies the connection between the subgraph $\ccalG_\kappa$ and the rest of the graph. Smaller $\beta_\ccalG(\ccalG_\kappa)$ is, weaker this connection becomes. The average relative subgraph degree $\bar{\beta}_{\ccalG} (\ccalG_0,\ldots,\ccalG_{q-1})$ generalizes this concept to the scenario with $q$ weakly connected subgraphs / clusters. The $\alpha$-realizable $q$-partition is a clustering property, which determines how well the graph is clustered. 
In particular, a low value of $\alpha$ indicates the graph has well-defined $q$ clusters and thus gives a topological characterization of the graph. With these preliminaries in place, we formally analyze how dropping intra-cluster edges affects graph topology in the following proposition. 
\begin{proposition}\label{proposition1:eigenvectorChange}
	Suppose a connected graph $\ccalG$ admits an $\alpha$-realizable $q$-partition [Def. \ref{def:alphaRelaization}] with clusters $\{\ccalG_0, \cdots, \ccalG_{q-1}\}$.
	Let $\ccalG'$ be the graph obtained from $\ccalG$ by dropping intra-cluster edges and $\{\ccalG'_0, \cdots, \ccalG'_{q-1}\}$ be the resulting clusters which is supposed to 
	be an $\alpha'$-realizable $q$-partition of $\ccalG'$. If $\{\mathbf{u}_0, \cdots, \mathbf{u}_{N-1}\}$ and  $\{\mathbf{u}'_0, \cdots, \mathbf{u}'_{N-1}\}$ are eigenvectors of the graph Laplacian of $\ccalG$ and $\ccalG'$ ordered by the increasing magnitude of eigenvalues, the distance between $\mathrm{span}\{\mathbf{u}_0, \mathbf{u}_1, \cdots, \mathbf{u}_{q-1}\}$ and $\mathrm{span}\{\mathbf{u}'_0, \mathbf{u}'_1, \cdots, \mathbf{u}'_{q-1}\}$ is bounded as
	\[ 
	\sqrt{ 
		\sum_{\kappa=0}^{q-1} \sum_{i=q}^{N-1} |\mathbf{u}_\kappa^{\mathsf{T}} \mathbf{u}'_i |^2 } ~~\leq~~ 
	\alpha + \alpha'.
	\]
\end{proposition}
\begin{proof}
	Let $\overline{\ccalG} = \cup_{\kappa=0}^{q-1} G_\kappa$ be the constituted graph of $q$ clusters $\{\ccalG_0, \cdots, \ccalG_{q-1}\}$ with the extreme clustering property [cf. \eqref{eq:constitutedGraph}], and $\{ \overline{\bbu}_0,\ldots,\overline{\bbu}_{N-1}\}$ be eigenvectors of the graph Laplacian of $\overline{\ccalG}$ (in order of increasing magnitude of corresponding eigenvalues). From Proposition $1$ in \cite{zhan2021barrier-journal} along with Def. \ref{def:relativeSubgraphDegree}--\ref{def:alphaRelaization}, the distance between $\mathrm{span}\{\mathbf{u}_0, \mathbf{u}_1, \cdots, \mathbf{u}_{q-1}\}$ and $\mathrm{span}\{ \overline{\bbu}_0,\ldots,\overline{\bbu}_{q-1}\}$ is bounded as
	\begin{align}\label{proofProp1:eq1}
		\sqrt{ \frac{1}{q(n-q)} 
			\sum_{\kappa=0}^{q-1} \sum_{i=q}^{N-1} |\overline{\mathbf{u}}_\kappa^{\mathsf{T}} \mathbf{u}_i |^2 } &\leq \frac{1}{\lambda_q} \sqrt{\frac{2}{N-q}} \bar{\beta}_{\ccalG}(\ccalG_0,\ldots,\ccalG_{q-1}) \\
		&\leq \frac{1}{\lambda_q} \sqrt{\frac{2}{N-q}} \frac{\alpha \lambda_q}{\sqrt{2q}} = \frac{\alpha }{\sqrt{q(N-q)}} \nonumber
	\end{align}
	where the second inequality is because $\ccalG$ admits an $\alpha$-realizable $q$-partition [cf. \eqref{eq:alphaRealizable}]. Thus, we have 
	$\sqrt{\sum_{\kappa=0}^{q-1} \sum_{i=q}^{N-1} |\overline{\mathbf{u}}_\kappa^{\mathsf{T}} \mathbf{u}_i |^2 } ~\leq~ \alpha $.
	Likewise, let ${\overline{\ccalG}}' = \cup_{j=0}^{q-1} G'_j$ be the constituted graph of $q$ clusters $\{\overline{\ccalG}_0, \cdots, \overline{\ccalG}_{q-1}\}$ after the intra-cluster edge dropping [cf. \eqref{eq:constitutedGraph}] and $\{\overline{\mathbf{u}}'_0, \overline{\mathbf{u}}'_1, \cdots, \overline{\mathbf{u}}'_{N-1}\}$ be eigenvectors of the graph Laplacian of ${\overline{\ccalG}}'$ (in order of increasing magnitude of corresponding eigenvalues). Since $\overline{\ccalG}'$ admits an $\alpha'$-realizable $q$-partition, we similarly have
	\begin{align}\label{proofProp1:eq15}
		\sqrt{\sum_{\kappa=0}^{q-1} \sum_{i=q}^{N-1} |{\overline{\mathbf{u}}'}_\kappa^{\mathsf{T}} \mathbf{u}'_i |^2 } ~\leq~ \alpha'
	\end{align}
	
	Note that the constituted graphs $\overline{\ccalG}$ and $\overline{\ccalG}'$ consist of $q$ disjoint clusters that are not connected to each other. Thus, the eigenvector $\overline{\bbu}_\kappa$ of $\overline{\ccalG}$ for $\kappa = 0, \ldots, q-1$ corresponds to a uniform positive distribution on the nodes of the cluster $\ccalG_\kappa$ and zero on all other nodes, so as the eigenvector $\overline{\bbu}_\kappa'$ of $\overline{\ccalG}'$ for $\kappa = 0, \ldots, q-1$. Since the clusters $\ccalG_\kappa$ and ${\ccalG_\kappa}'$ consist of the same nodes for $\kappa=0,\cdots,q-1$, the eigenvectors $\{\overline{\mathbf{u}}_\kappa\}_{\kappa=0,1,\cdots,q-1}$ and $\{\overline{\mathbf{u}}_\kappa'\}_{\kappa=0,1,\cdots,q-1}$ span the same vector subspace. Therefore, we have
	\begin{align}\label{proofProp1:eq2}
		\sqrt{\sum_{\kappa=0}^{q-1} \sum_{i=q}^{N-1} |{\overline{\mathbf{u}}}_\kappa^{\mathsf{T}} \overline{\mathbf{u}}'_i |^2 } ~=~ 0.
	\end{align}
	From Lemma 2 in \cite{bhattacharya2021geometric}, for any pair of orthonormal bases $\{\mathbf{w}_i\}_{i=0,1,\cdots,N-1}$ and $\{\mathbf{z}_i\}_{i=0,1,\cdots,N-1}$ of $\mathbb{R}^N$, the space distance
	\begin{align}\label{proofProp1:eq3}
		d_{\mathrm{sp}} (\mathrm{span}\{\mathbf{w}_0, ..., \mathbf{w}_{q-1}\}, \mathrm{span}\{\mathbf{z}_0, ..., \mathbf{z}_{q-1}\} ) := \sqrt{ \sum_{\kappa=0}^{q-1} \sum_{i=q}^{N-1} |\mathbf{w}_\kappa^{\mathsf{T}} \mathbf{z}_i |^2 }
	\end{align}
	is a true metric in the spaces of $q$-dimensional vector subspace of $\mathbb{R}^N$. Hence, it allows us to use the triangle inequality to complete the proof
	\begin{align}
		&\sqrt{ 
			\sum_{\kappa=0}^{q-1} \sum_{i=q}^{N-1} |\mathbf{u}_\kappa^{\mathsf{T}} \mathbf{u}'_i |^2 } = d_{\mathrm{sp}} (\mathrm{span}\{\mathbf{u}_0, ..., \mathbf{u}_{q-1}\}, \mathrm{span}\{\mathbf{u}'_0, ..., \mathbf{u}'_{q-1}\} ) \\
		&\le \!d_{\mathrm{sp}} (\mathrm{span}\{\mathbf{u}_0, ..., \mathbf{u}_{q-1}\}, \mathrm{span}\{ \overline{\bbu}_0,...,\overline{\bbu}_{q-1}\} ) \!+\! d_{\mathrm{sp}} (\mathrm{span}\{\mathbf{u}'_0, ..., \mathbf{u}'_{q-1}\}, \mathrm{span}\{ {\overline{\bbu}'}_0,...,{\overline{\bbu}'}_{q-1}\} ) \nonumber \\
		& + d_{\mathrm{sp}} (\mathrm{span}\{ \overline{\bbu}_0,...,\overline{\bbu}_{q-1}\}, \mathrm{span}\{ {\overline{\bbu}'}_0,...,{\overline{\bbu}'}_{q-1} ) \nonumber \\
		& = \sqrt{\sum_{\kappa=0}^{q-1} \sum_{i=q}^{N-1} |\overline{\mathbf{u}}_\kappa^{\mathsf{T}} \mathbf{u}_i |^2 } + \sqrt{\sum_{\kappa=0}^{q-1} \sum_{i=q}^{N-1} |{\overline{\mathbf{u}}'}_\kappa^{\mathsf{T}} \mathbf{u}'_i |^2 } + \sqrt{\sum_{\kappa=0}^{q-1} \sum_{i=q}^{N-1} |{\overline{\mathbf{u}}}_\kappa^{\mathsf{T}} \overline{\mathbf{u}}'_i |^2 } \le \alpha + \alpha'
	\end{align}
	where \eqref{proofProp1:eq1}, \eqref{proofProp1:eq15} and \eqref{proofProp1:eq2} are used in the last inequality. 
	%
\end{proof}
Proposition \ref{proposition1:eigenvectorChange} demonstrates that in a well-clustered graph, the distance between the subspaces spanned by the first $q$ eigenvectors (corresponding to the lowest $q$ eigenvalues) before and after the intra-cluster edge dropping is bounded by a measure of the clustering property within the graph. In essence, it states that dropping intra-cluster edges does not change significantly the first $q$ eigenvectors of the graph Laplacian as long as the graph maintains similarly well-clustered topology, where the first $q$ eigenvectors are typically utilized to characterize the structural information of the graph with $q$ clusters. It is worth noting that the bound on the perturbation of the eigenvector subspace 
can be arbitrarily small if $\alpha$ and $\alpha'$ are arbitrarily small, which can happen when large clusters are connected by weak inter-cluster edges even when a considerable number of intra-cluster edges are dropped. This result validates the fact that inter-cluster edges are more important than intra-cluster edges w.r.t. the topological connectivity in graphs with well-defined clusters.


\section{Method Details}

In this section, we provide more details with respect to the proposed TADropEdge method. 

\subsection{Disjoint component identification}

We consider three real-life datasets Cora, CiteSeer and Pubmed, and there is no new dataset included in experiments. For these datasets, the underlying graph is with a large number of nodes and edges such that it may contain disjoint components, where nodes in the same component can be connected either directly or
indirectly via intermediate nodes while nodes between different components can never find a connecting path. TADropEdge first preprocesses the underlying graph to identify these disjoint components. Table \ref{table:dataStatics} shows the statistics of three datasets and Table \ref{table:preprocessingResults} shows the preprocessing results. Specifically, we count the number of disjoint components, the maximal number of nodes in a single disjoint component (among all disjoint components), and the minimal number of nodes in a single disjoint componet (among all disjoint components).
\begin{table}[h]
	\begin{center}
		\caption{Dataset statistics.}
		\label{table:dataStatics}
		\begin{tabular}{|l|c|c|c|c|c|}
			\hline
			Dataset & Nodes & Edges & Classes & Features & Train / Validation / Testing  \\ \hline
			Cora &  2,708 & 5,429 & 7 & 1,433 & 1,207/500/1000  \\  \hline
			CiteSeer & 3,327 & 4,732 & 6 & 3,703 & 1812/500/1000 \\  \hline
			Pubmed & 19,717 & 44,338 & 3 & 500 & 18,217/500/1000 \\ 
			\hline
		\end{tabular}
	\end{center}  \vspace{-4mm}
\end{table}
\begin{table}[h]
	\begin{center}
		\caption{Preprocessing results for three datasets Cora, CiteSeer and Pubmed.}
		\label{table:preprocessingResults}
		\begin{tabular}{|l|c|c|c|}
			\hline
			Dataset & Disjoint components & Maximal number of nodes & Minimal number of nodes \\ \hline
			Cora &  78  & 2,484 & 2   \\  \hline
			CiteSeer & 814 & 2,120 & 1  \\  \hline
			Pubmed & 1 & 19,717 & 19,717  \\ 
			\hline
		\end{tabular}
	\end{center}  \vspace{-4mm}
\end{table}

\subsection{Edge weight computation}

\begin{figure*}[t]
	\begin{subfigure}{0.3\columnwidth}
		\includegraphics [width=0.9\linewidth, height = 0.85\linewidth]
		{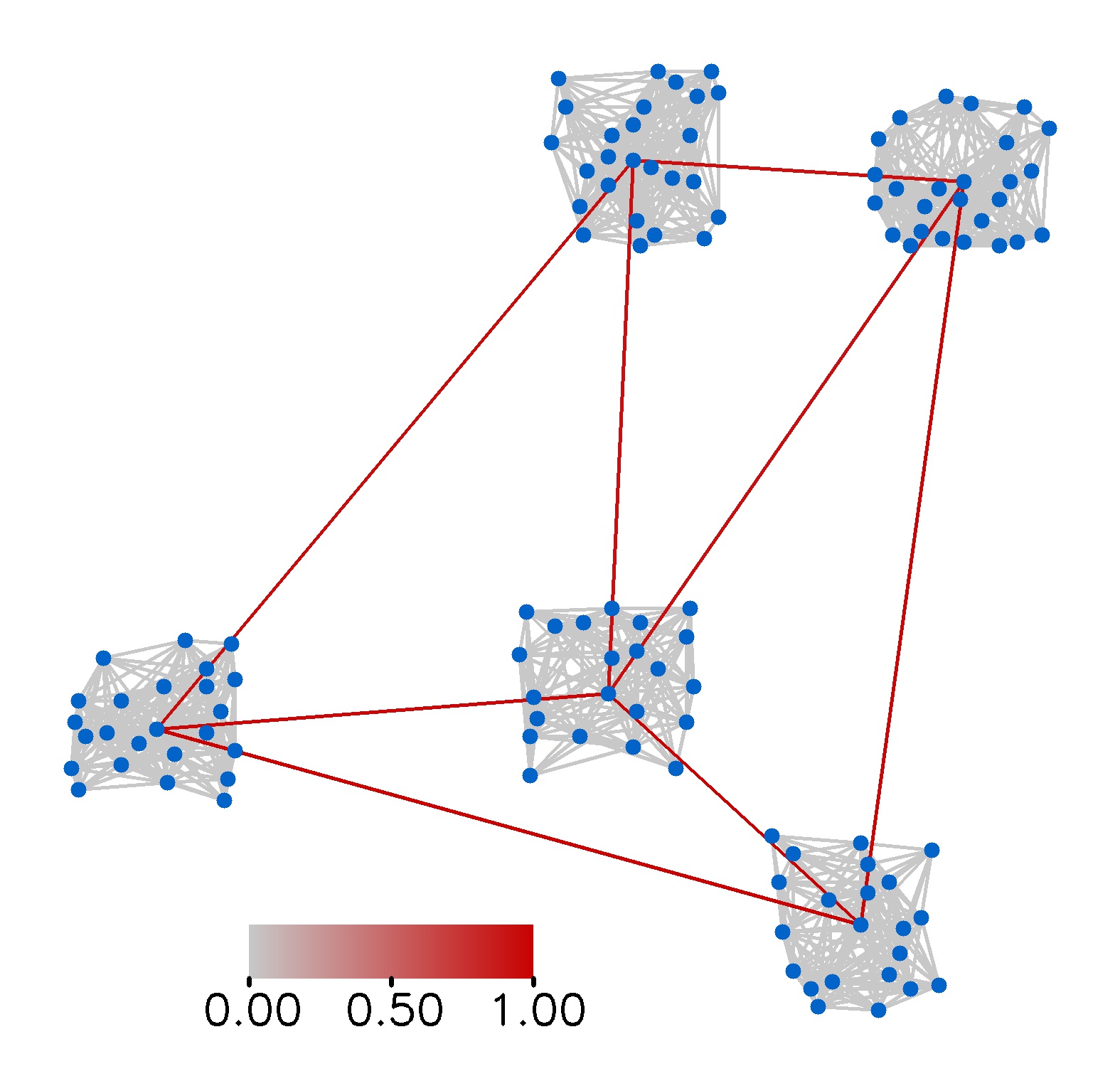}\qquad
		\caption{}%
		\label{subfig1_graph_example}%
	\end{subfigure}\hfill\hfill%
	\begin{subfigure}{0.3\columnwidth}
		\includegraphics [width=1.1\linewidth, height = 0.85\linewidth]
		{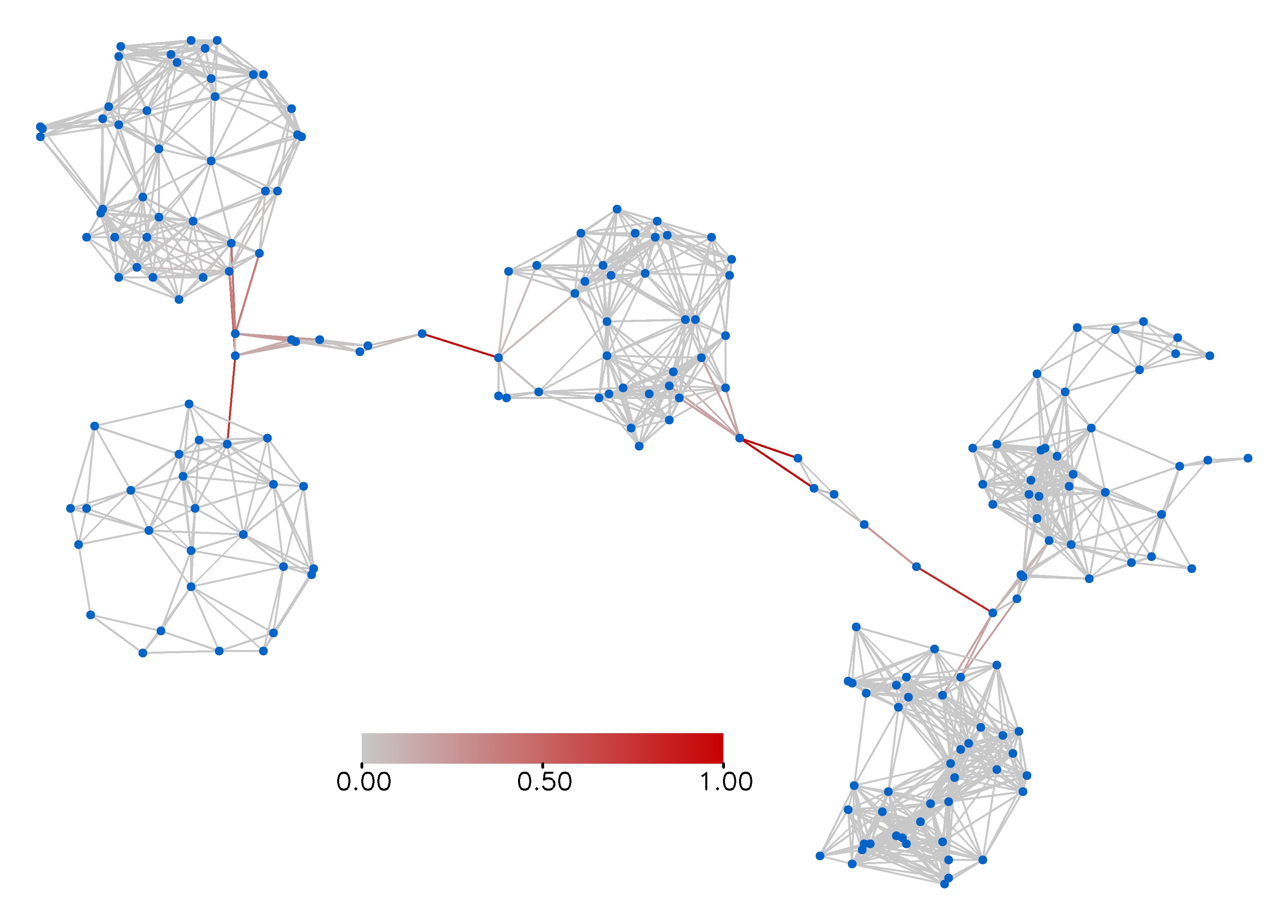}\qquad
		\caption{}%
		\label{subfig2_graph_example}%
	\end{subfigure}\hfill\hfill%
	\begin{subfigure}{0.3\columnwidth}
		\includegraphics [width=1.1\linewidth, height = 0.85\linewidth]
		{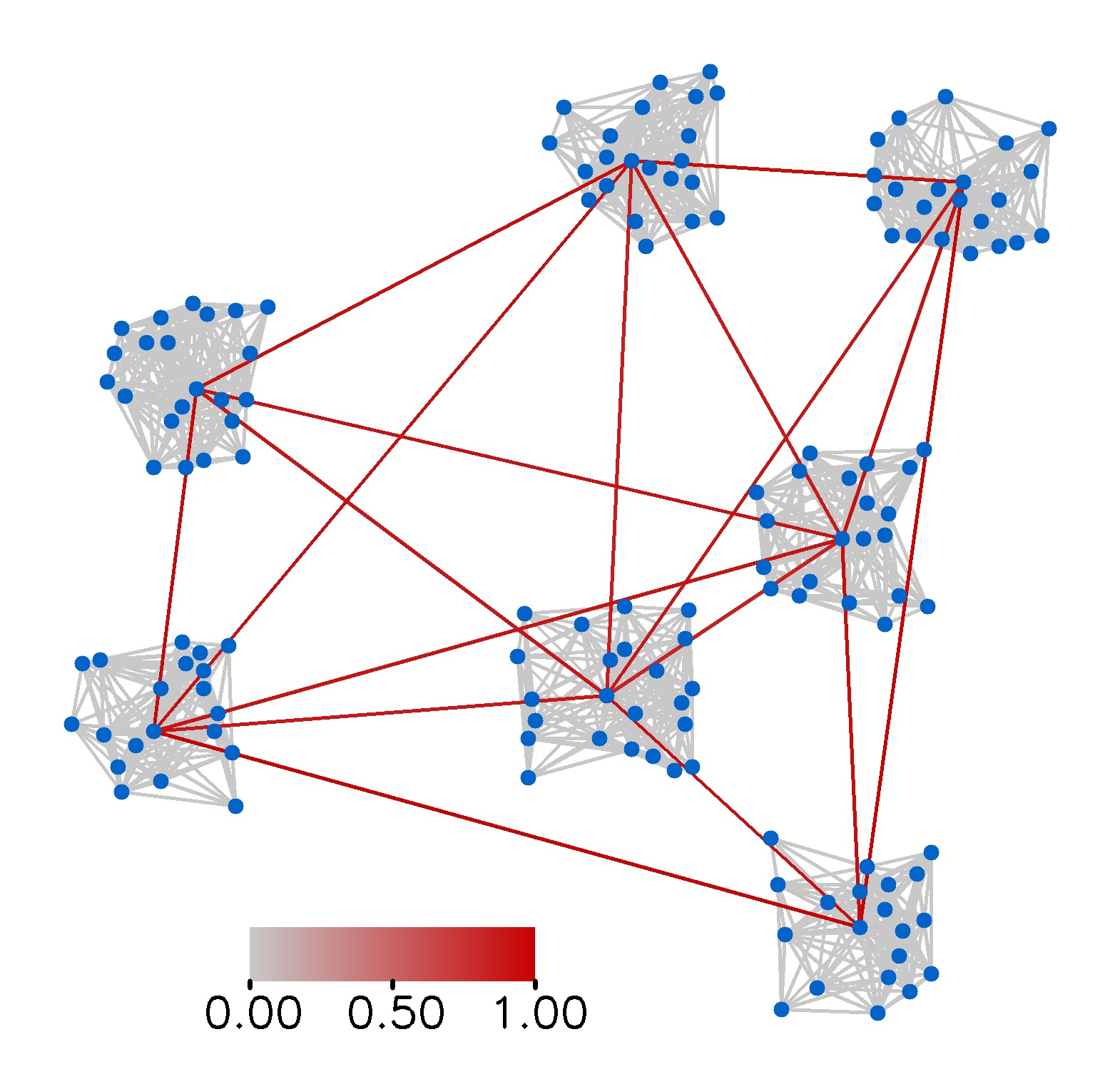}\qquad
		\caption{}%
		\label{subfig3_graph_example}%
	\end{subfigure}
	\caption{Aggregate resistance weights of three example graphs. (a) Graph $1$ with $3$ clusters. (b) Graph $2$ with $5$ clusters. (c) Graph $3$ with $7$ clusters.}
	\label{fig:resistanceWeightExamples}
\end{figure*}

We suppose there are $q$ clusters in each disjoint component (with no less than $q$ nodes), and compute aggregate resistance weights for each disjoint component separately [cf. (12) in the full paper]. The latter  
represent roles component edges play in being inter- or intra-cluster edges and thus imply their criticality to the component connectivity. During the weight computation among the disjoint components with at least $q$ nodes, the lowest weight is logged and set as the default value for all edges of the remaining disjoint components with less than $q$ nodes. 
To illustrate the effectiveness of aggregate resistance weights, besides Fig. 1a in the full paper, we provide three more examples in Fig. \ref{fig:resistanceWeightExamples} for single connected graphs with $5$, $5$ and $7$ clusters respectively. The edge color represents its corresponding weight value. We see that aggregate resistance weights successfully identify critical inter-cluster edges within the underlying graph, where larger weights imply more importance. 

\begin{figure*}[t]
	\begin{subfigure}{0.3\columnwidth}
		\includegraphics [width=1.1\linewidth, height = 0.85\linewidth]
		{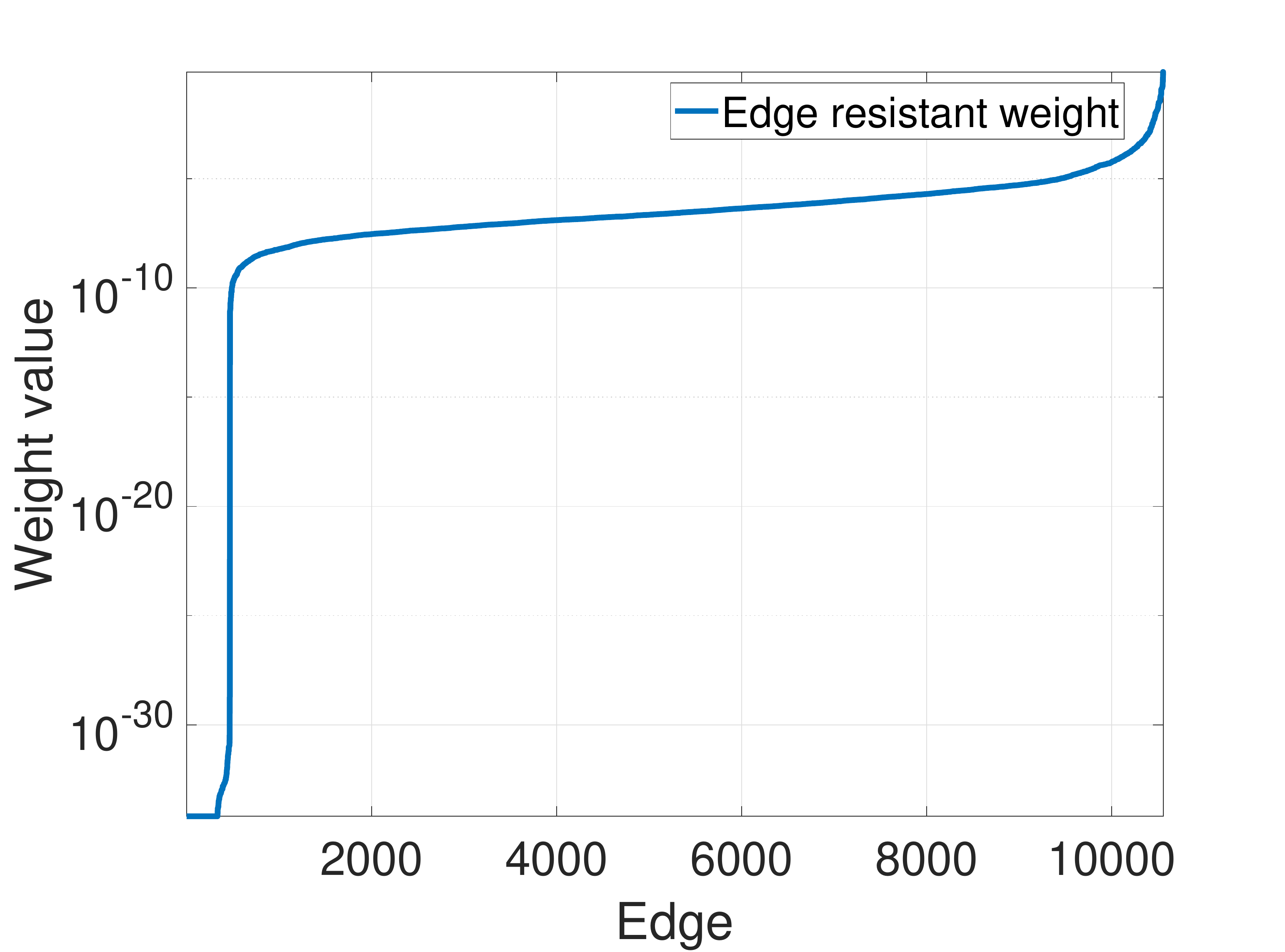}\qquad
		\caption{}%
		\label{subfig1_weight}%
	\end{subfigure}\hfill\hfill%
	\begin{subfigure}{0.3\columnwidth}
		\includegraphics [width=1.1\linewidth, height = 0.85\linewidth]
		{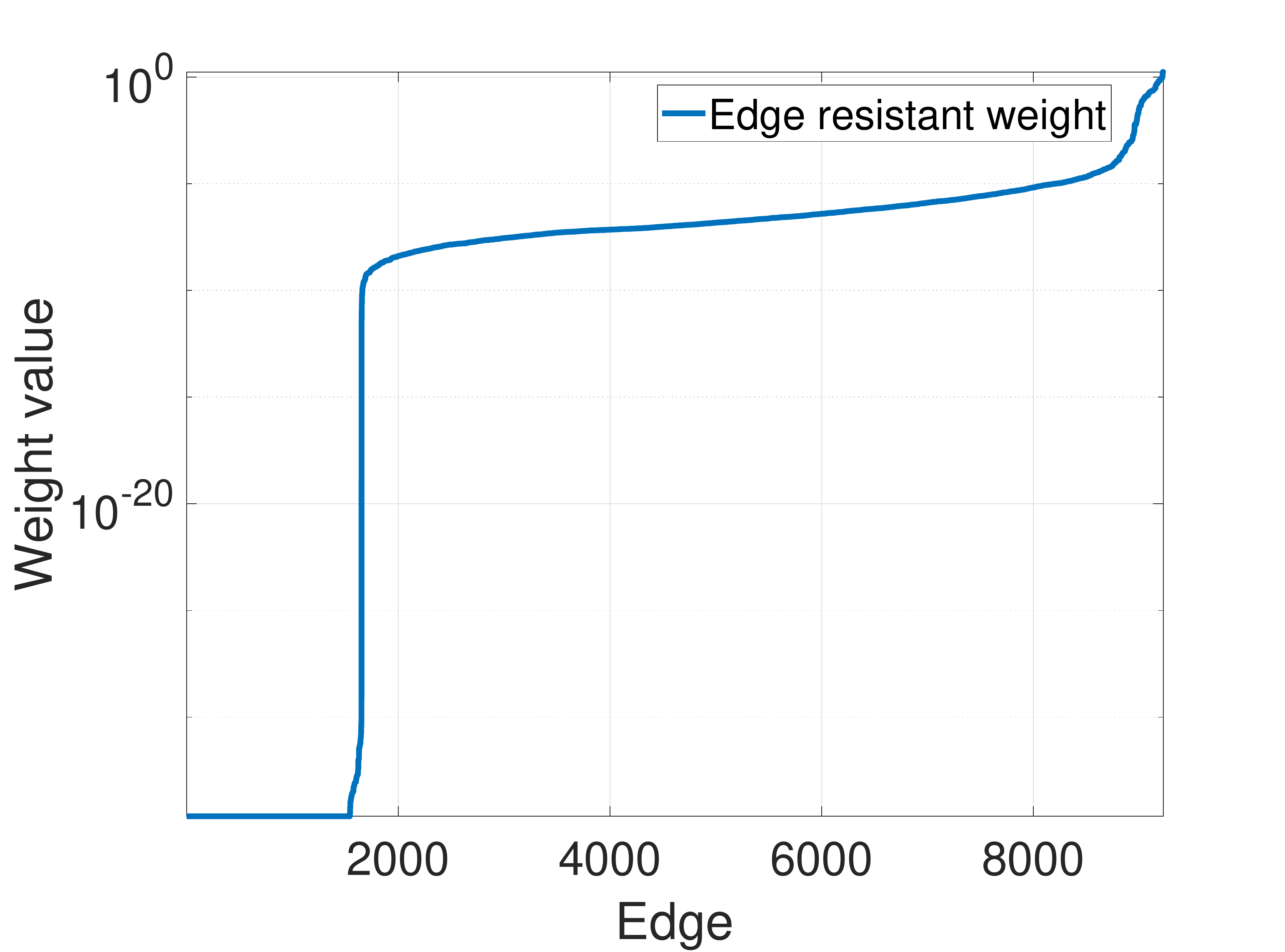}\qquad
		\caption{}%
		\label{subfig2_weight}%
	\end{subfigure}\hfill\hfill%
	\begin{subfigure}{0.3\columnwidth}
		\includegraphics [width=1.1\linewidth, height = 0.85\linewidth]
		{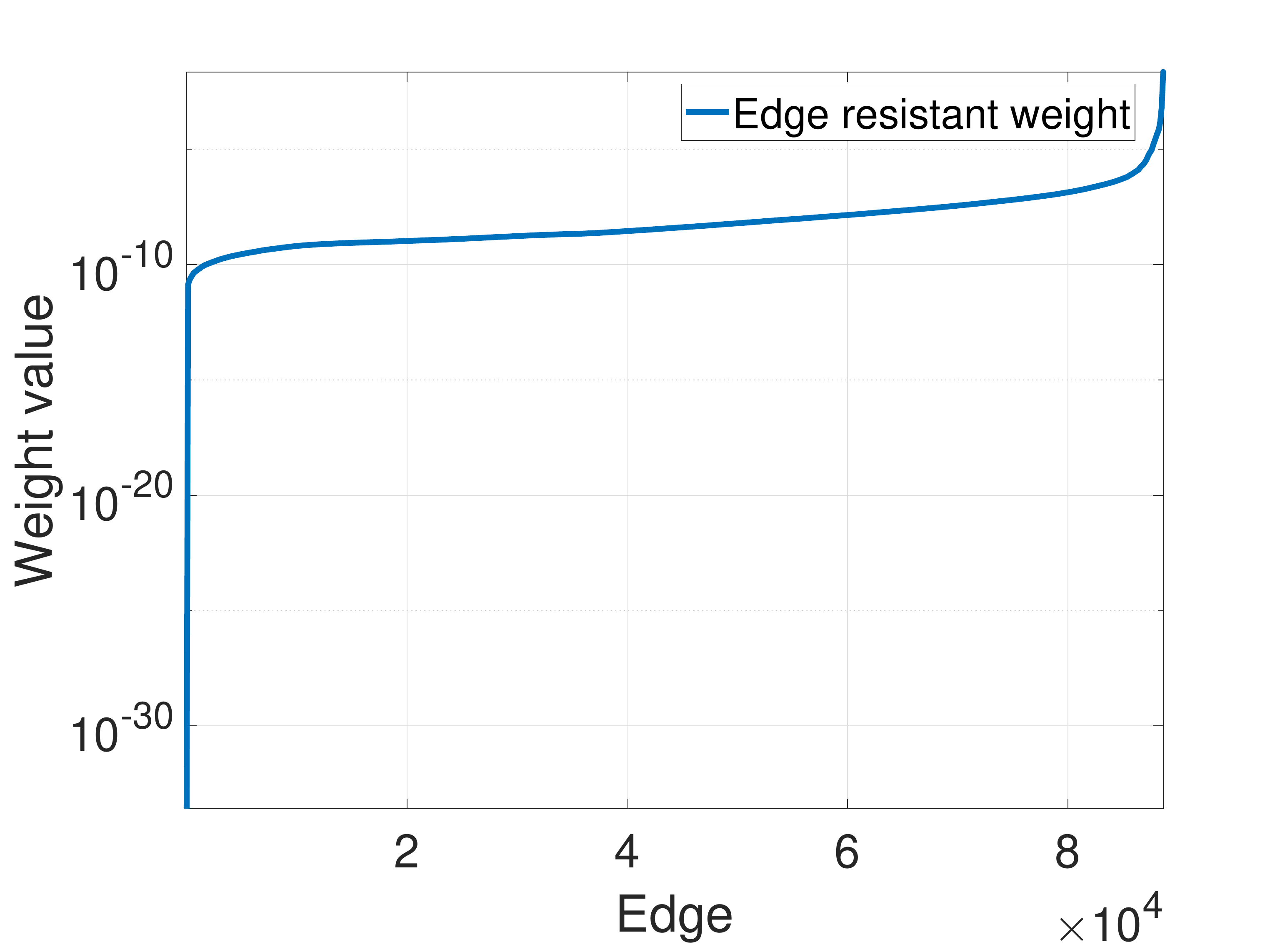}\qquad
		\caption{}%
		\label{subfig3_weight}%
	\end{subfigure}
	\caption{Aggregate resistance weight distribution over graph edges. (a) Cora. (b) CiteSeer. (c) Pubmed.}
	\label{fig:resistanceWeightDatasets}
\end{figure*}

For real-life datasets Cora, CiteSeer and Pubmed, Fig. \ref{fig:resistanceWeightDatasets} shows the aggregate resistance weight distribution over graph edges. Note that each edge $e_m$ with parent nodes $i_m$ and $j_m$ is considered with two pairs $(n_{i_m}, n_{j_m})$ and $(n_{j_m}, n_{i_m})$ in Fig. \ref{fig:resistanceWeightDatasets}, whose aggregate resistance weights are the same, i.e., $\omega(n_{i_m}, n_{j_m}) = \omega(n_{j_m}, n_{i_m})$. It is observed that the majority of edges have small weights as intra-cluster edges while a small number of edges have large weights as critical inter-cluster edges, as expected in theoretical analysis. 

\subsection{Topology adaptive edge dropping}\label{subsec:topologyAdaptive}

We propose three adaptive sampling strategies: \emph{threshold cutoff}, \emph{division normalization} and \emph{CDF normalization} for TADropEdge (Section 5 in the full paper). For each strategy, its parameter $\gamma$ is determined by the aggregate resistance weight distribution. In Fig. \ref{fig:resistanceWeightDatasets}, we observe a sudden increase of edge weights for a small number of edges in all three datasets and thus select the value of $\gamma$ from that interval. Specifically, the edge weight increases rapidly around the value $1 \cdot 10^{-2}$ in Cora, CiteSeer and the value $1 \cdot 10^{-3}$ in Pubmed. Therefore, we perform a random parameter search in the neighborhood of $1 \cdot 10^{-2}$ for Cora, CiteSeer and $1\cdot 10^{-3}$ for Pubmed to select an optimal $\gamma$ for the adaptive sampling strategy.

\section{Implementation Details}

We proceed to provide implementation details for both node-level and graph-level classifications.

\textbf{Hardware.} All experiments are implemented on a Windows machine with Intel(R) Core(TM) i7-8750H CPU (@ 2.20GHz) and 16GB RAM. The methods are accelerated by NVIDIA GeForce GTX1070 GPU with 16GB RAM.

\subsection{Node-level classification}\label{subsec:implementationNode}

\textbf{Backbones.} We consider three backbones: Graph Convolutional Network (GCN) \cite{Kipf2017}, Dense Network (JKNet) \cite{huang2018adaptive, Xu19-GIN} and Inception Network (IncepGCN) \cite{szegedy2016rethinking}. The architectures of these models can also be found in \cite{rong2019dropedge}, which are not repeated here to avoid content duplication.

\textbf{Hyper-parameter optimization.} We follow the experimental setting in \cite{rong2019dropedge} for fair comparison. Specifically, we adopt Adam optimizer to train each model for $400$ epochs. To ensure the re-productivity of the results, the seeds of the random numbers of all experiments are set to the same.

We set the hidden dimension as $128$ and perfrom a random search strategy for the other hyper-parameters, which are summarized in Table \ref{table:hyperParameters}. For each model, we try $100$ different hyper-parameter combinations via random search and select the best test accuracy as the result, which is reported in Table 1 of the full paper. Regarding the same architecture with TADropEdge or with DropEdge or without edge dropping, we apply the same set of hyper-parameters for fair comparison.
\begin{table}[h]
	\begin{center}
		\caption{Hyper-parameter description.}
		\label{table:hyperParameters}
		\begin{tabular}{|l|l|}
			\hline
			Hyper-parameter & Description \\ \hline
			lr &  Learning rate   \\  \hline
			weight-decay & L2 regulation weight  \\  \hline
			dropout &  Dropout rate   \\  \hline
			$p$ & Default sampling probability  \\  \hline
			$\gamma$ & Adaptive sampling strategy parameter  \\ 
			\hline
		\end{tabular}
	\end{center}  \vspace{-4mm}
\end{table}

\subsection{Graph-level classification}


\textbf{Dataset.} We consider the signal diffusion process over the stochastic block model (SBM) graph of $N=50$ nodes equally divided into $C=5$ communities. There exists a source node $n_s$ in each community for $s\in \{s_1,\ldots,s_5\}$. The initial source signal is a Kronecker delta $\bbdelta_s = [\delta_1, \ldots, \delta_N]^\top \in \{ 0,1 \}^N$ with $\delta_s = 1$ at the source node $s \in \{ s_1, \ldots, s_5 \}$ and $\delta_i = 0$ at the other nodes $i \neq s$. The diffused signal at time $t$ is $\bbx_{st} = \bbS^t \bbdelta_s + \bbn$ where $\bbS \in \mathbb{R}^{N \times N}$ is the normalized adjacency matrix and $\bbn \in \mathbb{R}^{N}$ is the additional noise whose components are drawn from the normal distribution $\ccalN(0, 2.5\cdot 10^{-2})$. The training dataset consists of $1600$ signal-label samples $\{ (\bbx_{st}, n_s) \}$ by randomly picking a source node $s \in \{ s_1, \ldots, s_5 \}$, a diffused time $t \in \{ 1, \ldots, 50 \}$ and an additional noise $\bbn$, which is split into $100$ samples for training, $500$ samples for validation and $1000$ samples for testing.

\textbf{Backbone.} We consider a single-layered Graph Convolutional Neural Network (GCNN) \cite{Fernando2019, Wu19-SGC}, which contains $F=32$ features of filter order $K=5$ [cf. (1) in the full paper] and ReLU nonlinearity in the layer. We adopt Adam optimizer for training and similarly perform a random parameter search to report the best result for each method (i.e., TADropEdge, DropEdge and the original model).

\section{Inverse-TADropEdge}

\begin{figure*}[t]
	\begin{subfigure}{0.3\columnwidth}
		\includegraphics [width=1.1\linewidth, height = 0.85\linewidth]
		{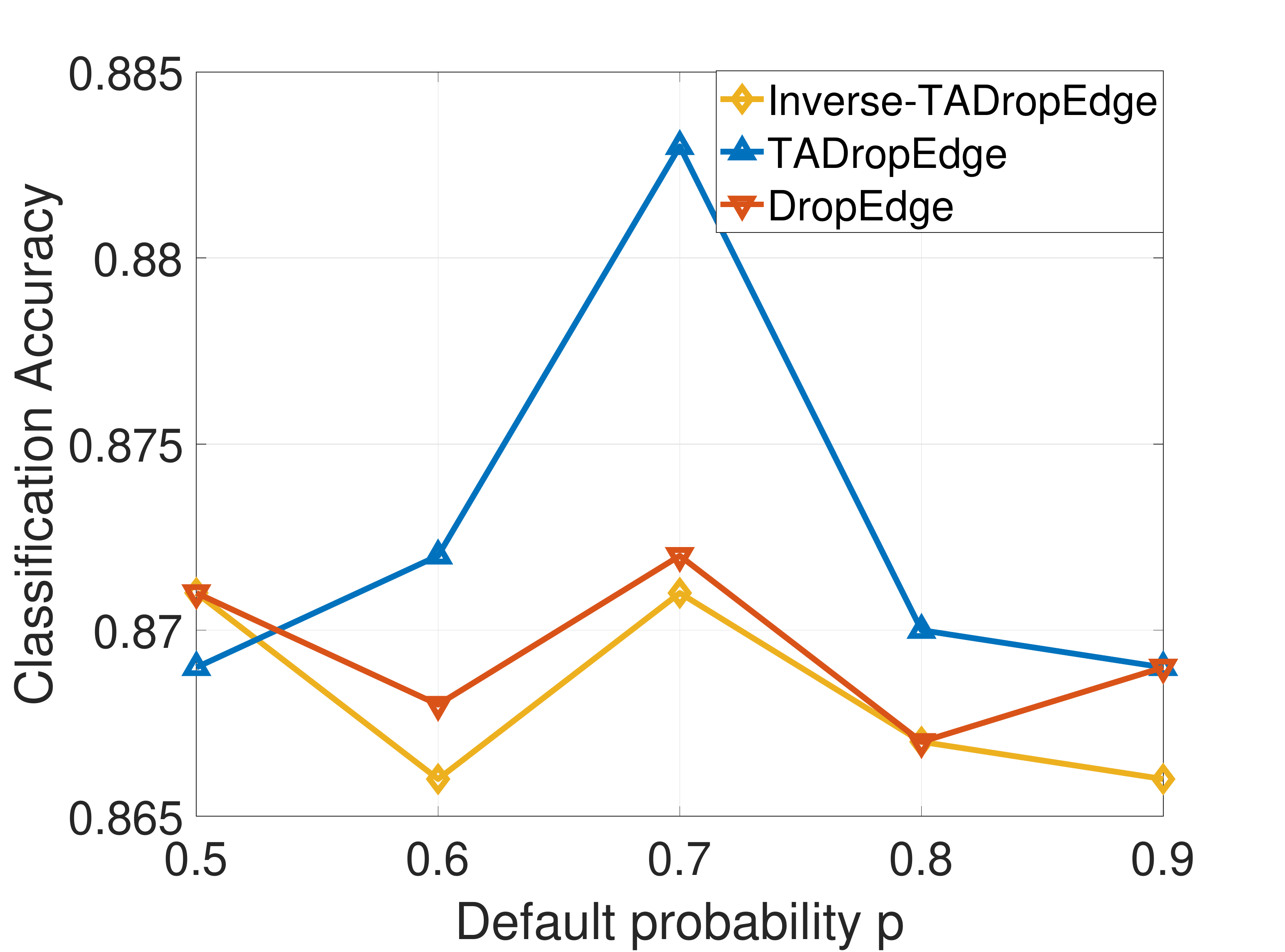}\qquad
		\caption{}%
		\label{subfig1_inverse}%
	\end{subfigure}\hfill\hfill%
	\begin{subfigure}{0.3\columnwidth}
		\includegraphics [width=1.1\linewidth, height = 0.85\linewidth]
		{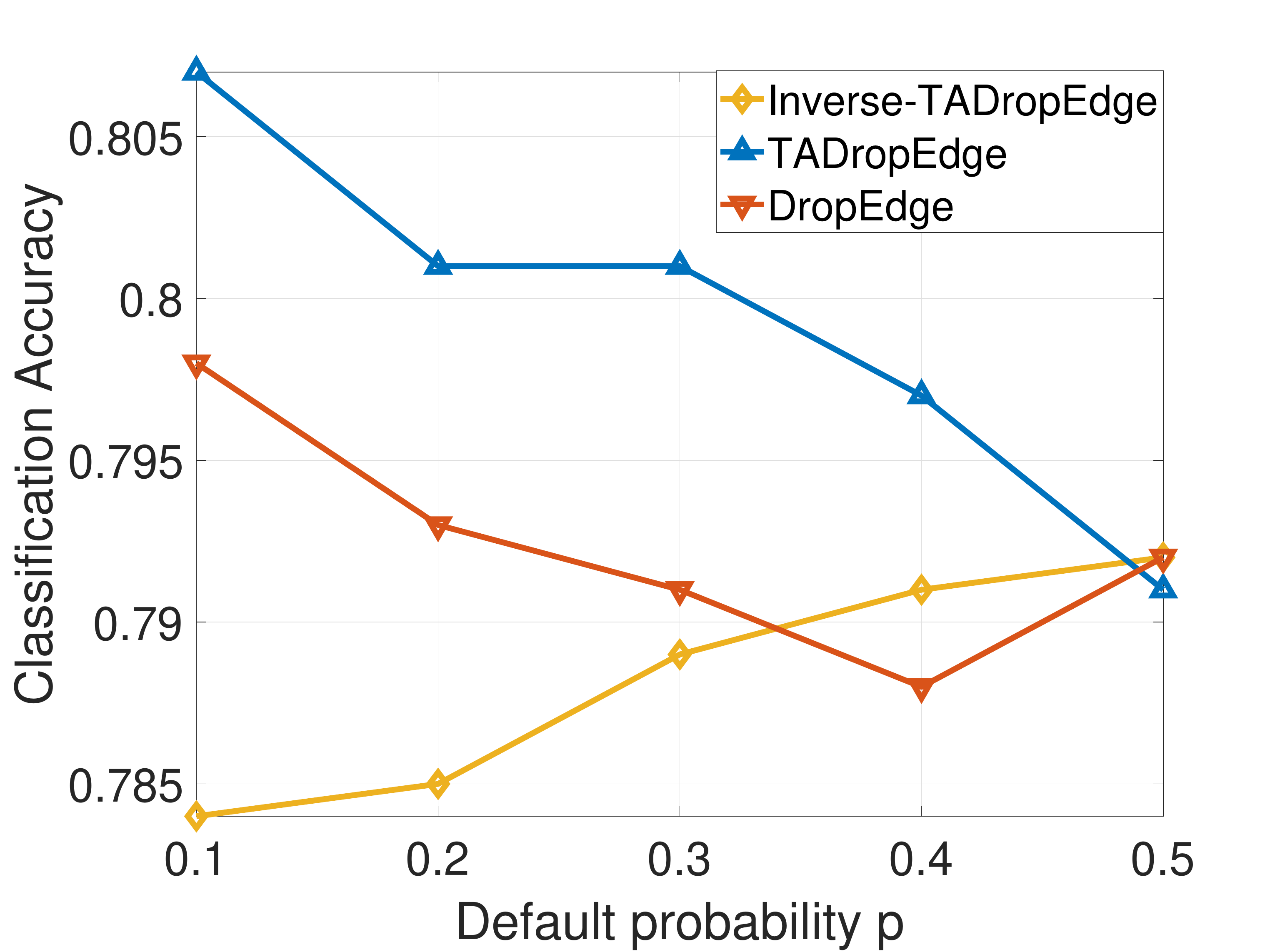}\qquad
		\caption{}%
		\label{subfig2_inverse}%
	\end{subfigure}\hfill\hfill%
	\begin{subfigure}{0.3\columnwidth}
		\includegraphics [width=1.1\linewidth, height = 0.85\linewidth]
		{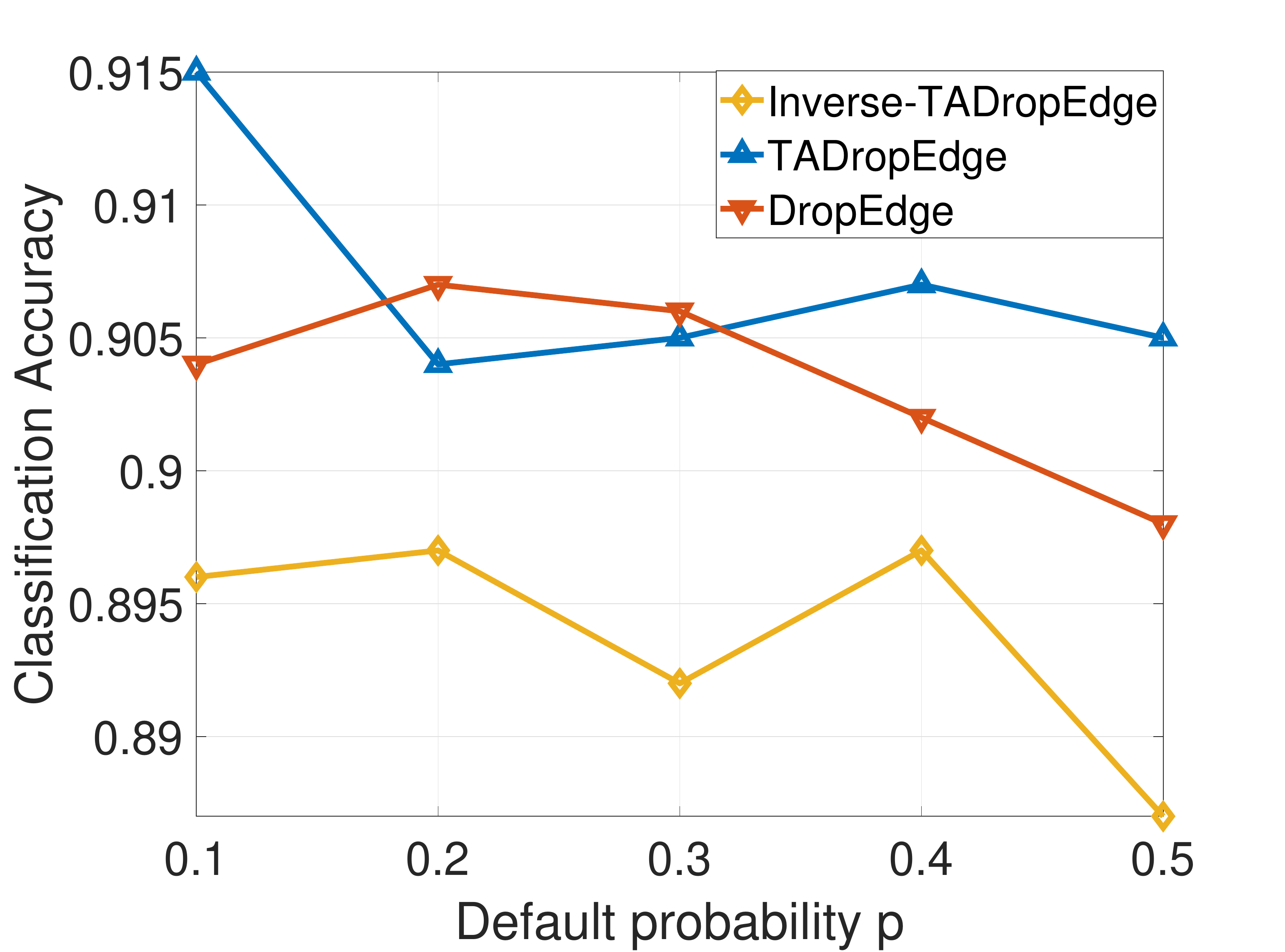}\qquad
		\caption{}%
		\label{subfig3_inverse}%
	\end{subfigure}
	\caption{Performance comparison between Inverse-TADropEdge, TADropEdge and DropEdge. (a) Cora. (b) CiteSeer. (c) Pubmed.}
	\label{fig:InverseTADropEdge}
\end{figure*}

To further evaluate the effectiveness of the proposed method, we consider an inverse version of TADropEdge, referred to as Inverse-TADropEdge, which performs in an opposite manner to TADropEdge. In particular, Inverse-TADropEdge samples intra-cluster edges at higher probabilities while sampling inter-cluster edges at lower probabilities. It thus samples edge-dropped subgraphs that break the overall connectivity of the underlying graph and carry little structural information inherent in graph signals, resulting in noisy information during training and degrading performance of data augmentation. In this context, we expect that Inverse-TADropEdge would perform worse than not only TADropEdge but also i.i.d. DropEdge.

We modify three adaptive sampling strategies for Inverse-TADropEdge as follows:

(i) \text{\emph{Inverse threshold cutoff}:} We consider a threshold $\gamma$ determined by the edge weight distribution. Graph edges are sampled at the default probability $p$ if their weights are \textbf{larger} (---\emph{for TADropEdge, here is "\textbf{smaller}"}) than $\gamma$, otherwise they are maintained as undropped. It drops a small number of edges that are critical for maintaining graph connectivity.

(ii) \text{\emph{Inverse division normalization}:} We normalize the edge weights to $[0,1]$ with the division function, and determine the sampling matrix $\bbP_{inv,\ccalG,p}$ by the normalized weights. Given the edge weight $\omega$ and the function parameter $\gamma$, the edge sampling probability is $~p_{inv,\ccalG,p} = p + (1-p)*\gamma/(\gamma + \omega)$ (---\emph{for TADropEdge, here is "\textbf{$~p_{\ccalG,p} = 1 - (1-p)*\gamma/(\gamma + \omega)$}"}).

(iii) \text{ \emph{Inverse CDF normalization}:} We normalize the edge weights to $[0,1]$ with the cumulative distribution function (CDF), and determine the sampling matrix $\bbP_{inv,\ccalG,p}$ by the normalized weights. Given the edge weight $\omega$ and the CDF $f(\omega)$ of $\omega$, the edge sampling probability is $~p_{inv,\ccalG, p} = 1 - (1-p) * f(\omega)$ (---\emph{for TADropEdge, here is "$~p_{\ccalG, p} = p + (1-p) * f(\omega)$"}). 

In contrast to TADropEdge, all strategies of Inverse-TADropEdge sample graph edges of higher weights at lower probabilities close to $p$ while sampling graph edges of lower weights at higher probabilities up to $1$, where the parameter $\gamma$ is selected based on the edge weight distribution as well (Section \ref{subsec:topologyAdaptive}). 
At each training epoch $t$, Inverse-TADropEdge samples an edge-dropped subgraph $\ccalG_t$ with the sampling matrix $\bbP_{inv,\ccalG,p}$, and replaces the shift operator $\bbS$ of the underlying graph $\ccalG$ with the sparse shift operator $\bbS_t$ of the subgraph $\ccalG_t$ in the architecture [cf. (2) in the full paper] for signal propagation and parameter training. 

We perform Inverse-TADropEdge on the $3$-layered GCN for three datasets Cora, CiteSeer and Pubmed. Note that this section mainly focuses on analyzing comparisons between Inverse-TADropEdge, TADropEdge and DropEdge without the concern of pushing state-of-the-art results, such that we do not perform delicate hyper-parameter selection. The random seed is fixed and other experimental settings follow Section \ref{subsec:implementationNode}. 
We select the CDF normalization as the adaptive sampling strategy for Inverse-TADropEdge and TADropEdge, which outperforms the other two strategies in more cases as observed in Table 1 of the full paper. 
Fig. \ref{fig:InverseTADropEdge} shows the comparison results under varying default probabilities $p$. We see that Inverse-TADropEdge exhibits the worst performance compared to TADropEdge and DropEdge in all three datasets. This is because Inverse-TADropEdge breaks the overall topology of the underlying graph, which results in edge-dropped subgraphs that carry little structural information embedded in graph signals and make noisy information that hurts the training process. This result further validates the effectiveness of TADropEdge, i.e., it emphasizes the importance of maintaining graph connectivity during random edge dropping when considering DropEdge as a data augmentation technique.

\section{Variance Reduction}

\begin{figure*}[t]
	\begin{subfigure}{0.3\columnwidth}
		\includegraphics [width=1.1\linewidth, height = 0.85\linewidth]
		{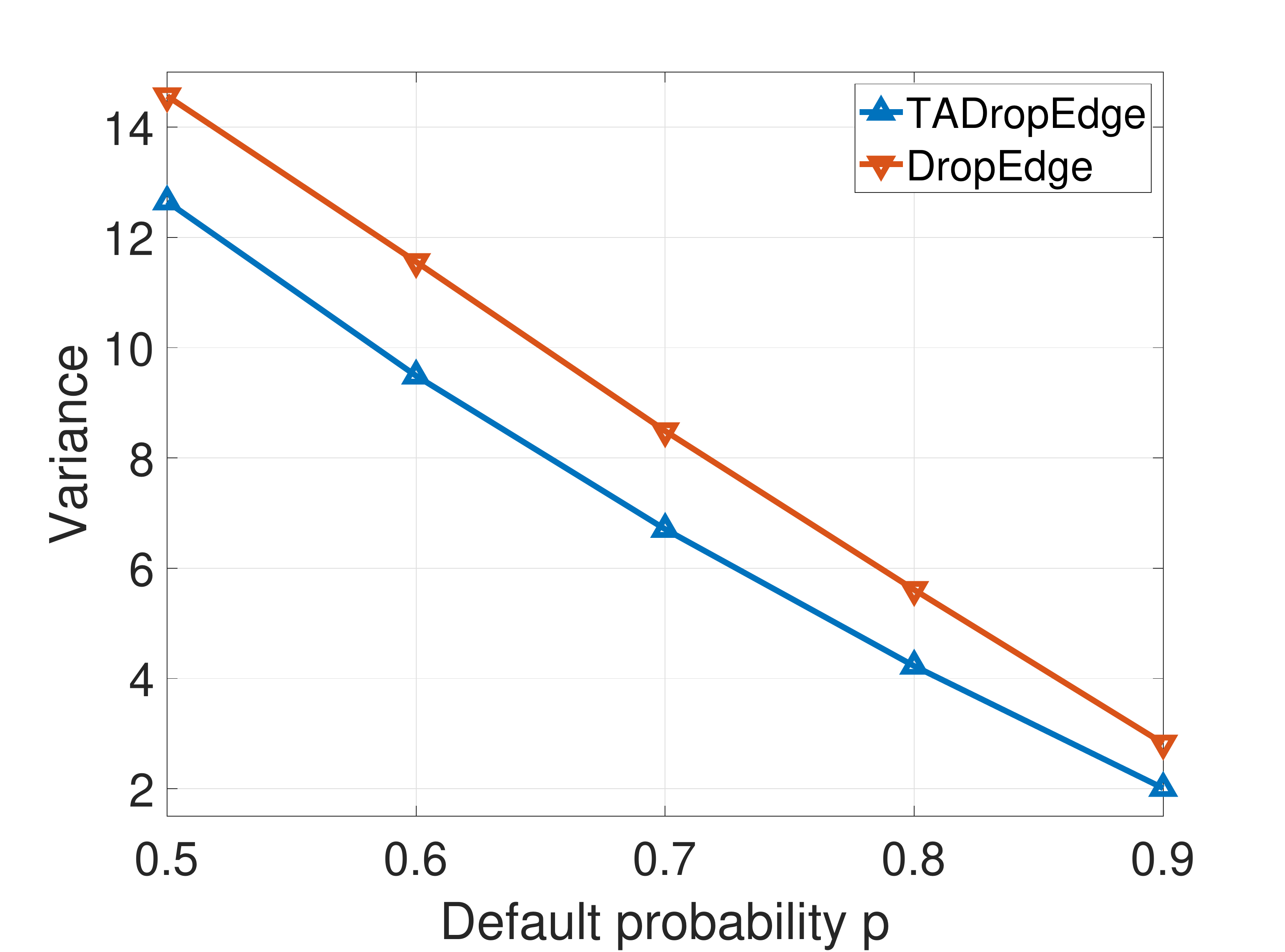}\qquad
		\caption{}%
		\label{subfig1}%
	\end{subfigure}\hfill\hfill%
	\begin{subfigure}{0.3\columnwidth}
		\includegraphics [width=1.1\linewidth, height = 0.85\linewidth]
		{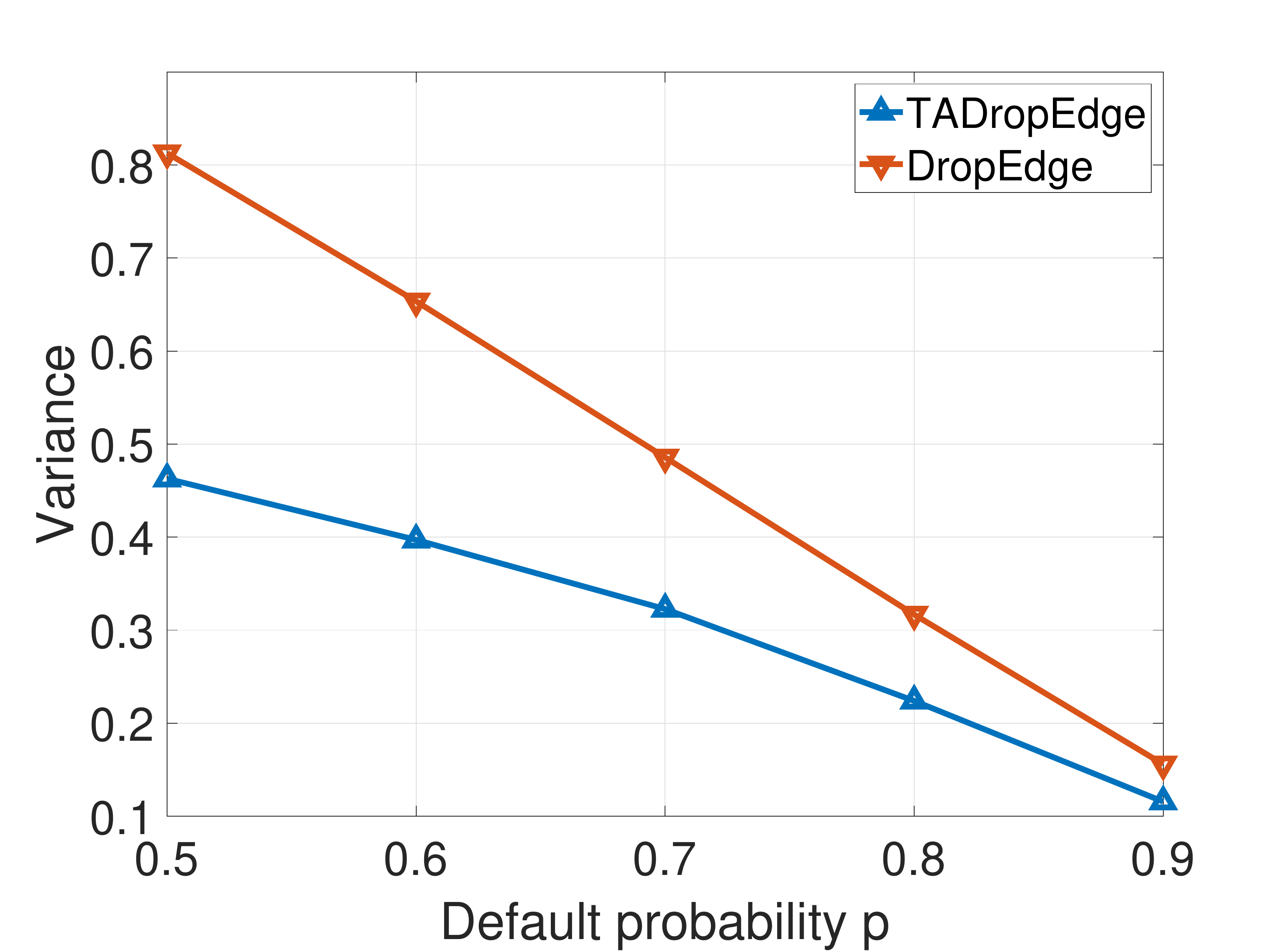}\qquad
		\caption{}%
		\label{subfig2}%
	\end{subfigure}\hfill\hfill%
	\begin{subfigure}{0.3\columnwidth}
		\includegraphics [width=1.1\linewidth, height = 0.85\linewidth]
		{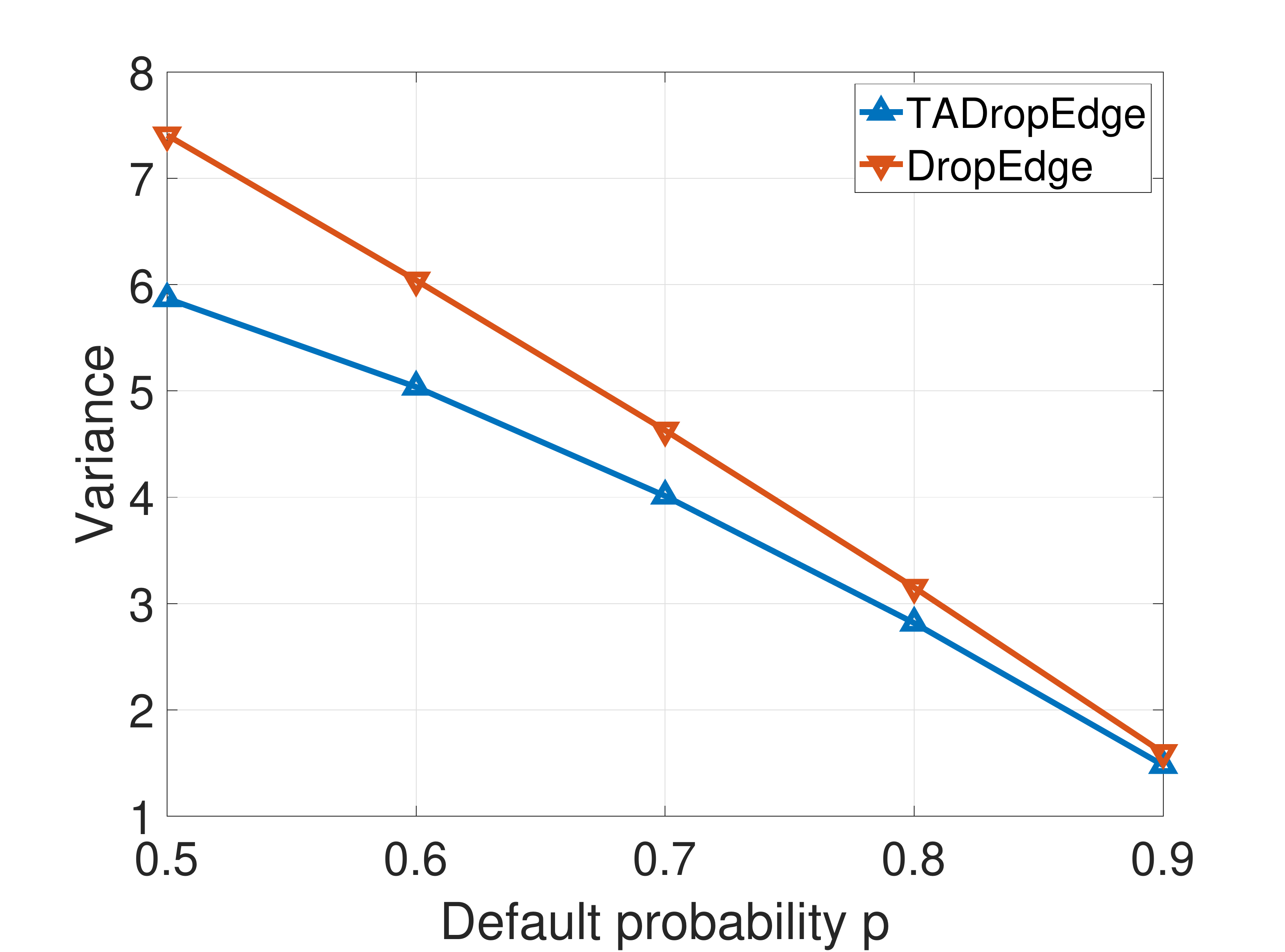}\qquad
		\caption{}%
		\label{subfig3}%
	\end{subfigure}
	\caption{Variance comparison between TADropEdge and DropEdge. (a) Cora. (b) CiteSeer. (c) Pubmed.}
	\label{fig:varianceComparison}
\end{figure*}

As discussed in Section 5.1 of the full paper, TADropEdge has reduced variance by sampling a small number of critical edges with higher probabilities or keeping them undropped. Fig. 2(a) in the full paper corroborates this theoretical finding by presenting converging behaviors of TADropEdge and DropEdge. In this section, we further validate this aspect by directly comparing the variance of output features between TADropEdge and DropEdge. We consider the $3$-layered GCN on three datasets Cora, CiteSeer and Pubmed. The architecture parameters are randomly initialized and no training is performed. 
For our analysis, we consider the variance over all graph nodes as
\begin{align}\label{eq:variance}
	\text{var}[\bbPhi(\bbX; \bbS, \ccalA)] = \sum_{i=1}^N \frac{1}{F_L} \sum_{j=1}^{F_L} \text{var}\big[[\bbPhi(\bbX; \bbS, \ccalA)]_{ij}\big]
\end{align}
where $[\bbPhi(\bbX; \bbS, \ccalA)]_{ij}$ is the $(i,j)$th entry of the $L$th layer (final layer) output feature $\bbPhi(\bbX; \bbS, \ccalA) \in \mathbb{R}^{N \times F_L}$, representing the $j$th feature at the $i$th node. Note that $\text{var}[\cdot]$ in \eqref{eq:variance} is with respect to random edge-dropped subgraphs.

Fig. \ref{fig:varianceComparison} shows the results, where the CDF mormalization is selected as the adaptive sampling strategy in TADropEdge. The variance increases as the default probability $p$ decreases with more graph randomness involved. Though TADropEdge only maintains a small number of inter-cluster edges, it reduces variance significantly on all three datasets. The variance reduction increases with the decreasing of $p$, which is especially remarkable on CiteSeer achieving maximal $43\%$ reduction. These results emphasize the role played by inter-cluster edges in maintaining the overall topology of the underlying graph; hence, GCNs built upon edge-dropped subgraphs yield different but similar output features as the GCN built upon the underlying graph. The reduced variance obtained by TADropEdge then accelerates the training process and mitigates the training difficulty.



\end{document}

%% file: mySections.tex
\usepackage{needspace}

